\documentclass[letterpaper]{article} 
\usepackage[preprint]{aaai2027}
\usepackage[hyphens]{url}  
\usepackage{graphicx} 
\usepackage{natbib}  
\usepackage{caption} 
\usepackage[utf8]{inputenc}
\usepackage{soul}
\usepackage{subcaption}
\usepackage{pifont}
\usepackage{amsmath}
\usepackage{amsthm}
\usepackage{booktabs}
\usepackage{algorithm}
\usepackage{algorithmic}
\usepackage{pgfplots}
\usepackage{threeparttable}
\usepackage{multirow}
\usepackage{chemformula}

\usepackage{amsmath,amsfonts,bm}

\def\eqref#1{equation~\ref{#1}}

\def\floor#1{\lfloor #1 \rfloor}
\def\1{\bm{1}}

\def\vd{{\bm{d}}}
\def\ve{{\bm{e}}}

\def\vg{{\bm{g}}}
\def\vh{{\bm{h}}}

\def\vm{{\bm{m}}}

\def\vo{{\bm{o}}}

\def\vr{{\bm{r}}}
\def\vs{{\bm{s}}}

\def\vy{{\bm{y}}}

\def\mC{{\bm{C}}}

\def\mE{{\bm{E}}}

\def\mG{{\bm{G}}}
\def\mH{{\bm{H}}}

\def\mS{{\bm{S}}}

\def\mX{{\bm{X}}}

\def\mZ{{\bm{Z}}}

\def\mPhi{{\bm{\Phi}}}

\DeclareMathAlphabet{\mathsfit}{\encodingdefault}{\sfdefault}{m}{sl}
\SetMathAlphabet{\mathsfit}{bold}{\encodingdefault}{\sfdefault}{bx}{n}

\def\gE{{\mathcal{E}}}

\def\gG{{\mathcal{G}}}

\def\gL{{\mathcal{L}}}
\def\gM{{\mathcal{M}}}
\def\gN{{\mathcal{N}}}

\def\gV{{\mathcal{V}}}

\def\sR{{\mathbb{R}}}

\newcommand{\E}{\mathbb{E}}

\pgfplotsset{compat=1.18}
\definecolor{cCDVAE}{HTML}{FF7F0E}
\definecolor{cMatterGen}{HTML}{2CA02C}
\definecolor{cGraphite}{HTML}{9467BD}
\definecolor{cODE}{HTML}{8C564B}
\definecolor{cSDE}{HTML}{E377C2}
\definecolor{cShortcut}{HTML}{D62728}

\pgfplotsset{
  scatteraxis/.style={
    width=8.4cm, height=5.6cm,
    grid=major, grid style={gray!25, line width=0.3pt},
    tick align=outside,
    tick label style={font=\scriptsize},
    label style={font=\small},
    legend cell align=left,
    legend style={font=\fontsize{5.5}{6.5}\selectfont, draw=black,
                  fill=white, fill opacity=0.9, text opacity=1,
                  inner sep=2pt, row sep=-1pt},
  },
  scpts/.style={
    only marks, mark size=2.4pt,
    nodes near coords, point meta=explicit symbolic,
    every node near coord/.append style={
      font=\fontsize{4}{4.5}\selectfont, inner sep=0pt, outer sep=1pt,
      anchor=south west, /pgf/number format/assume math mode=true},
  },
  sCDVAE/.style={color=cCDVAE, mark=triangle*,
    mark options={rotate=180, fill opacity=0.85},
    every node near coord/.append style={text=cCDVAE}},
  sMatterGen/.style={color=cMatterGen, mark=pentagon*,
    mark options={fill opacity=0.85},
    every node near coord/.append style={text=cMatterGen}},
  sGraphite/.style={color=cGraphite, mark=triangle*,
    mark options={fill opacity=0.85},
    every node near coord/.append style={text=cGraphite}},
  sODE/.style={color=cODE, mark=*,
    mark options={fill opacity=0.85},
    every node near coord/.append style={text=cODE}},
  sSDE/.style={color=cSDE, mark=square*,
    mark options={fill opacity=0.85},
    every node near coord/.append style={text=cSDE}},
  sShortcut/.style={color=cShortcut, mark=diamond*,
    mark options={fill opacity=0.85},
    every node near coord/.append style={text=cShortcut}},
}

\newtheorem{definition}{Definition}
\newcommand{\archname}[0]{\textsc{AMShortcut}}

\newcommand{\scoreX}{\nabla_\mX\log p(\mX_t)}
\newcommand{\scoreE}{\nabla_\mE\log p(\mE_t)}

\title{\archname: An Inference- and Training-Efficient Inverse Design Model for Amorphous Materials}

\author{
    Yan Lin\textsuperscript{\rm 1}\equalcontrib,
    Jonas A. Finkler\textsuperscript{\rm 1}\equalcontrib,
    Tao Du\textsuperscript{\rm 2},
    Jilin Hu\textsuperscript{\rm 3},
    Morten M. Smedskjaer\textsuperscript{\rm 1}
}
\affiliations{
    \textsuperscript{\rm 1}Aalborg University\\
    \textsuperscript{\rm 2}The Hong Kong Polytechnic University\\
    \textsuperscript{\rm 3}East China Normal University\\
    lyan@cs.aau.dk, jaf@bio.aau.dk, tao.du@polyu.edu.hk, hujilin@cs.aau.dk, mos@bio.aau.dk
}

\begin{document}

\maketitle

\begin{abstract}
Amorphous (disordered) materials are solids that lack long-range atomic order but possess complex short- and medium-range order, and require simulation cells with at least hundreds to thousands of atoms to accurately capture their diverse local atomic environments.
To advance the design of amorphous materials with desired properties, inverse design with probabilistic generative models has emerged as a promising approach, which directly generates the atomic positions and elements of amorphous materials given a set of desired properties.
In this paper, we focus on the computational efficiency of such models, introducing \archname, an inference- and training-efficient probabilistic generative model for amorphous materials. \archname{} enables accurate inference of amorphous materials with only a few sampling steps, mitigating the need for an excessive number of sampling steps that hinder inference efficiency. \archname{} can be trained once with all relevant properties and perform inference conditioned on different combinations of desired properties, mitigating the need for training one model for each combination that lowers training efficiency.
Its denoiser uses an efficient E(n)-equivariant backbone that captures the complex short- and medium-range order of amorphous materials at a low computational cost, making both inference and training more efficient.
Experiments on three amorphous material datasets with diverse structures and properties demonstrate that \archname{} achieves its design goals.
\end{abstract}

\section{Introduction}
Glasses and other amorphous materials are solids that lack a periodic atomic arrangement or long-range atomic order, yet exhibit complex short- and medium-range order.
Unlike crystalline materials, whose periodic structure allows them to be fully described by a small unit cell containing only a few to hundreds of atoms, amorphous materials require larger simulation cells with at least hundreds or thousands of atoms to accurately capture their diverse local atomic environments.
In other words, their atoms are randomly arranged overall but still form organized clusters in localized regions, i.e., clusters that cannot be represented by simple repetition of a basic unit.
They show great potential in energy storage, thermal management, and advanced materials~\citep{liu2024amorphous}.

To advance the design of amorphous materials with desired properties, instead of relying on resource-intensive trial-and-error processes~\citep{liu2017materials,cai2020machine}, inverse design has emerged as a promising approach.
It starts with target properties and works backward to determine the necessary atomic configurations. One intuitive way to implement this approach is through probabilistic generative models~\citep{DBLP:journals/corr/KingmaW13,goodfellow2014generative}, especially those based on diffusion models~\citep{DBLP:conf/nips/HoJA20}, which generate atomic positions and elements conditioned on desired properties by transforming random noise to targets through a multi-step Markov process.
Such models have shown success in generating crystalline materials and molecules~\citep{DBLP:conf/nips/WuGL0022,DBLP:conf/iclr/XieFGBJ22,DBLP:conf/icml/HoogeboomSVW22,zeni2025generative}, both of which can be represented with relatively few atoms as discussed above.
However, they remain underdeveloped for amorphous materials due to the lack of large-scale datasets and significant computational efficiency challenges when generating samples with hundreds to thousands of atoms~\citep{yang2025generative}.
Among these recent efforts, AMDEN~\citep{finkler2025inverse} is the state of the art for inverse design of amorphous materials. It requires many sampling steps for accurate generation and trains a separate model for each set of target properties, limiting its inference and training efficiency.

\begin{figure}[t]
    \centering
    \includegraphics[width=1.0\linewidth]{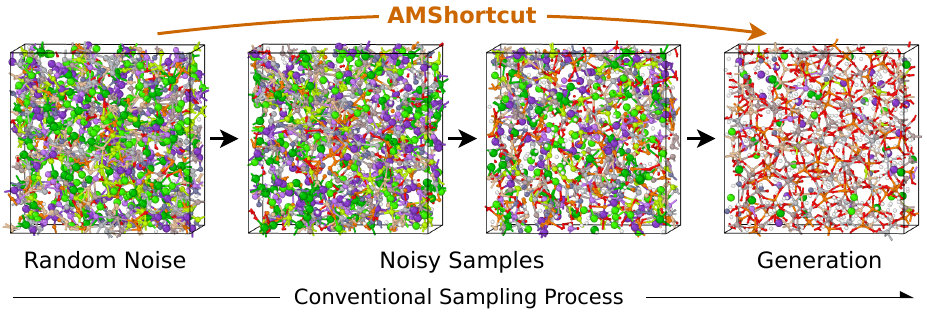}
    \caption{\archname{} generates structurally accurate amorphous material samples with one or few sampling steps, enabling high-throughput inverse design of amorphous materials.}
    \label{fig:framework}
\end{figure}

In this paper, we focus on the \textbf{inference and training efficiency} of probabilistic generative models for amorphous materials.
Specifically, \textit{inference efficiency} is hindered by both the large system size of amorphous material samples and the need for many sampling steps. As noted earlier, amorphous materials require large simulation cells often with hundreds to thousands of atoms to capture their short- and medium-range orders. This, combined with the complexity of exploring optimal atomic configurations in the absence of periodic structure, poses significant computational challenges.
We show that accurate generation of such structures indeed requires many sampling steps for state-of-the-art generative modeling frameworks including score-matching SDEs~\citep{DBLP:conf/iclr/0011SKKEP21} and flow-matching ODEs~\citep{DBLP:conf/iclr/LipmanCBNL23} (see Section~\ref{sec:structural-accuracy}).
On the other hand, \textit{training efficiency} is hindered by the variety of properties on which the generative models need to be conditioned. In practice, inverse design of amorphous materials focuses on different sets of properties to fit different needs, yet training one model for each set of properties requires training and maintaining numerous models. Techniques such as classifier guidance~\citep{DBLP:conf/nips/DhariwalN21,DBLP:conf/iclr/Lin0ZB25} enable training of one uniform unconditioned generative model and guide the model's generation with dedicated classifiers. Yet, the poor availability of differentiable classifiers for amorphous materials can make the implementation of such techniques impractical.

Here, we propose \textbf{\archname} (Figure~\ref{fig:framework}), a model for taking \underline{shortcuts} in \underline{a}morphous \underline{m}aterial design that serves as both an effective inverse design model and an efficient probabilistic generative model for amorphous materials.
It improves both the inference and training efficiency of generative modeling for amorphous materials, as we develop in the methodology.
Our contributions are as follows:
\begin{itemize}
\item We propose \archname, to our knowledge the first few-step generative model for inverse design of amorphous materials, built on a \textit{material differential equation framework} that unifies score-matching and flow-matching with ghost-atom density control.
\item We introduce a \textit{shortcut} for the diffusion-based generation of amorphous materials, whose self-consistency objective acts within our material differential equation framework jointly on the continuous atomic positions under periodic boundary conditions and the categorical elements. This enables accurate generation with few sampling steps, significantly improving inference efficiency.
\item We introduce a \textit{flexible material denoiser} that is trained once on all properties and can perform inference conditioned on different subsets of them, which avoids a separate model per property subset and improves training efficiency. Because this conditioning stays accurate with few sampling steps, the denoiser's training efficiency and \archname's inference efficiency hold together.
\item We design an \textit{efficient E(n)-equivariant backbone} for the material denoiser, a shallow and narrow message-passing network with an angular gate for short-range order and virtual context nodes for global context, which captures the short- and medium-range order of amorphous materials at a low computational cost and thereby improves the inference and training efficiency of \archname.
\item We conduct experiments on three amorphous material datasets, showing that with no more than ten sampling steps \archname{} closely matches the structural accuracy and inverse design performance of the strongest baseline, AMDEN, at 250 sampling steps, while reducing generation time by up to 99\%.
\end{itemize}

\section{Related Work}
\paragraph{Generative modeling of atomic systems.}
Recent years have seen significant efforts on generative modeling of molecules and crystalline materials. Early methods based on variational autoencoders~\citep{DBLP:journals/corr/KingmaW13,gebauer2019symmetry,noh2019inverse,court20203} are limited in effectively generating complex atomic systems~\citep{DBLP:conf/iclr/DaunhawerSCPV22}. Other methods based on generative adversarial networks (GANs)~\citep{goodfellow2014generative,long2021constrained} are hampered by the unstable training of GANs~\citep{li2018limitations}. More recent works based on diffusion models~\citep{DBLP:conf/nips/HoJA20,DBLP:conf/nips/WuGL0022,DBLP:conf/iclr/XieFGBJ22,DBLP:conf/icml/HoogeboomSVW22,zeni2025generative} have shown promising results.
Despite these efforts, molecules and crystalline materials are inherently distinct from amorphous materials. As discussed in the Introduction, molecules are usually not composed of many atoms, and crystalline materials feature strong periodic atomic order that enables them to be represented by relatively few atoms in a periodic cell.
In contrast, amorphous materials lack long-range atomic order and require large simulation cells with thousands of atoms to accurately capture their diverse local environments. This fundamental difference in representation size poses significant computational efficiency challenges for existing generative models.

\paragraph{Inverse design of amorphous materials.}
\citet{zhou2023generative} introduces a generative framework for predicting compositions of glass materials given desired properties. However, compositions do not provide a complete picture of atomic configurations of glasses and do not fully determine their properties (e.g., the thermal history also matters).
There are a few efforts on generating atomic configurations of amorphous materials, some based on GAN~\citep{xu2023generative,yong2024dismai} and others based on VAE~\citep{chen2025physical,kilgour2020generating,li2025conditional}. As mentioned before, their generation quality is limited by the shortcomings of the underlying GAN and VAE frameworks.
Diffusion models~\citep{DBLP:conf/nips/HoJA20} are used to generate structures of amorphous carbon with desired spectroscopy~\citep{kwon2024spectroscopy}, yet their exploration of broader properties and more diverse multi-element systems is limited.
Recent works expand diffusion model-based inverse design to more diverse amorphous materials~\citep{yang2025generative,finkler2025inverse}, yet the challenges of subpar inference and training efficiency remain unsolved.
Overall, the inverse design of amorphous materials remains underdeveloped.

\section{Preliminary}

\begin{definition}[Amorphous material sample]
An amorphous material sample is represented as the positions and elements of atoms inside a periodic cell, formally a tuple $\gM=(\mC, \mX, \mE)$ of three matrices.
$\mC \in \sR^{3\times 3}$ contains the three lattice vectors of the cell, $\mX \in \sR^{n_a\times 3}$ represents the positions of $n_a$ atoms, and $\mE \in \sR^{n_a\times d_E}$ contains the one-hot embeddings of atomic elements, where $d_E$ is the total number of elements under consideration.
The set of $n_p$ relevant properties is represented as $\vy\in \sR^{n_p}$, where each value denotes the magnitude of that property.
\end{definition}

\begin{definition}[Ghost atoms]
We introduce ghost atoms, a special atom type incorporated into each material sample, to allow the generative model to control the density of the sample without modifying $\mC$ or the number of atoms.
In the datasets, ghost atoms are randomly positioned in the cell so that the total number of atoms $n_a=\floor{\rho\cdot\text{Vol}(\mC)}$, where $\rho$ is the maximum density.
Ghost atoms are treated like normal atoms by the model but are assigned a special chemical element class. The model adjusts sample density by changing the fraction of ghost atoms, removed after generation.
\end{definition}

\noindent \textbf{Problem definition.}
\textit{Inverse design of amorphous materials.}
Given a set of desired properties $\vy$, the goal of inverse design is to generate amorphous material samples $\gM$ that exhibit these properties. Formally, we aim to sample from the conditional distribution $p(\gM|\vy)$, generating new samples with the specified target properties.

\section{\archname{} Model}

\subsection{Material Differential Equation}
Our goal is to sample from the distribution of samples conditioned on properties, i.e., $p(\gM|\vy)$, to generate new samples with desired properties. Since the exact form of this distribution is unknown, we parameterize the sampling process as a time-dependent differential equation, which we term \textit{material differential equation}:
\begin{equation}
   d\gM_t=\mu(\gM_t,\vy,t)dt+\sigma(t)dW_t\quad t\in[1, 0],
   \label{eq:material-diff-eq}
\end{equation}
which defines a continuous-time process that transforms a noise sample $\gM_1$ that can be easily sampled from a prior distribution to a target sample $\gM_0$.
$\mu(\gM_t,\vy,t)$ is the deterministic drift coefficient and $\sigma(t)$ is the diffusion coefficient controlling the magnitude of the stochastic Wiener process $W_t$.
In practice, the sampling process is applied to positions $\mX$ and element embeddings $\mE$ with the cell $\mC$ unchanged, and both $\mu$ and $\sigma$ have two components for positions and element embeddings respectively, which we denote as $\mu_{\mX}$, $\mu_{\mE}$, $\sigma_{\mX}$, and $\sigma_{\mE}$.

To generate a sample $\gM_0$, we sample noise $\gM_1$ and solve the above differential equation using the Euler-Maruyama method, where we discretize the time span $[1, 0]$ into $n_s$ steps with step size $\Delta t=1/n_s$. Each step is performed as:
\begin{equation}
   \gM_{t-\Delta t}=\gM_t - \Delta t \mu(\gM_t,\vy,t) + \sqrt{\Delta t}\sigma(t)\epsilon\quad \epsilon\sim\gN(0,1)
   \label{eq:euler}
\end{equation}
This can be intuitively understood as moving the atomic positions and element embeddings of a sample in the direction and speed given by $\mu(\gM_t,\vy,t)$, while adding a certain magnitude of noise.
In practice, the drift coefficient $\mu$ is estimated by a learnable neural network $\mu_\theta(\gM_t,\vy,t)$ with parameters $\theta$, while the diffusion coefficient $\sigma$ is predefined.
The network is trained by sampling $\gM_0$ from the dataset and the corresponding $\gM_t$ from a forward process, then regressing $\mu_\theta$ on the ground truth drift coefficient $\mu$.

\subsubsection{Material ODE and Material SDE}
We instantiate the material differential equation with two variants that differ in their forward process and ground truth.
The \textit{material ODE} is the deterministic variant with $\sigma(t)=0$, following the optimal transport flow~\citep{DBLP:conf/iclr/LipmanCBNL23}.
The \textit{material SDE} is the stochastic variant following the score-matching SDE~\citep{DBLP:conf/iclr/0011SKKEP21}, with a variance exploding noise schedule on positions and a variance preserving noise schedule with cosine progression on element embeddings.
In both variants, inference samples the positions $\mX_1$ from a uniform distribution within the cell $\mC$ and the element embeddings $\mE_1$ from a standard normal distribution, while training obtains $\gM_t$ from the ground truth $\gM_0$ through the forward process.
The detailed formulations are given in Appendix~\ref{apx:material-ode} and \ref{apx:material-sde}.

Material ODE has no stochastic component and is more stable for generation with a small number $n_s$ of steps, while the stochastic component of material SDE gives it more freedom to explore structures during generation.
In practice both variants still require many steps to generate structurally accurate samples (Section~\ref{sec:structural-accuracy}), which motivates the shortcut training introduced next.

\subsection{Learning Shortcuts in Material SDE}
Works on one-step diffusion models~\citep{DBLP:conf/iclr/FransHLA25,geng2025mean} provide insights into why material ODE and SDE require many sampling steps. The direction and speed defined by the drift coefficient $\mu$ change rapidly over time $t$. In Eq.~\ref{eq:euler}, the material sample is updated using the instantaneous drift at the current time, resulting in inaccurate generation when a large step size is used.
Inspired by these works, the proposed \archname{} is built on material SDE, but has additional awareness of the step size and provides \textit{shortcuts}, i.e., instead of updating the sample with the instantaneous drift, it updates the sample accurately across long step sizes.

Specifically, \archname{} trains a neural network $u_\theta(\gM_t,\vy,t,\Delta t)$ with similar architecture to $\mu_\theta$ but additionally takes the step size $\Delta t$ into consideration. A ground truth shortcut $u$ across $t$ and $t-\Delta t$ is defined as:
\begin{equation}
\begin{split}
  u(\gM_t, \vy, t, \Delta t) =&
  \frac{1}{\Delta t} \int_{t-\Delta t}^{t} \mu(\gM_\tau,\vy,\tau) d\tau \\
  &+ \frac{1}{\Delta t} \int_{t-\Delta t}^{t} \sigma(\tau) dW_\tau
\end{split}
\end{equation}
Note that similar to $\mu$, a shortcut also contains two components for positions and element embeddings denoted as $u_\mX$ and $u_\mE$ respectively, but we do not present the separate formulas for simplicity. The network $u_\theta$ still predicts the noise first, and the predicted shortcuts are calculated following the parameterization of $\mu$ in material SDE.

To learn the shortcuts in a computationally efficient way, we follow the idea of self-consistency loss in \citet{DBLP:conf/iclr/FransHLA25} and adapt it to our setting, where the shortcut loss is defined within the material differential equation framework and jointly handles the continuous atomic positions $\mX$ and the categorical element embeddings $\mE$:
\begin{equation}
   \begin{aligned}
   \gL_\text{SC} &= \E_{\gM_0,t,\Delta t} \| \text{pbc}( \gM_t - 2\Delta t\, u_\theta(\gM_t,\vy,t,2\Delta t) \\
   &\quad - \text{sg}(\gM_\text{target}) ) \|^2 \\
   \gM_\text{target} &= \hat{\gM}_{t-\Delta t} - \Delta t\, u_\theta(\hat{\gM}_{t-\Delta t},\vy,t-\Delta t,\Delta t) \\
   \hat{\gM}_{t-\Delta t} &= \gM_t - \Delta t\, u_\theta(\gM_t,\vy,t,\Delta t),
   \end{aligned}
   \label{eq:shortcut-loss}
\end{equation}
where $\gM_\text{target}$ is the sample reached by two shortcut steps of size $\Delta t$ through the intermediate $\hat{\gM}_{t-\Delta t}$, $\text{sg}$ is stop gradient, and $\text{pbc}(\cdot)$ takes the position difference along its shortest path across the periodic boundaries of $\mC$.
We use the fact that two consecutive shortcuts reach the same sample as one shortcut with double the step size, which $\text{pbc}(\cdot)$ keeps consistent when an atom crosses a cell boundary during the large jumps of few-step generation.
For more stable training, the steps in Eq.~\ref{eq:shortcut-loss} follow \citet{DBLP:conf/iclr/0011SKKEP21} in removing the stochastic component of material SDE and modifying the drift coefficients in compensation.
Detailed formulations are given in Appendix~\ref{apx:mdshortcut}.

Finally, $u_\theta$ is trained with the combined loss of material SDE and shortcuts, $\gL=\gL_\text{SDE}+\gL_\text{SC}$, using the equivalence $u_\theta(\gM_t,\vy,t,0)\equiv \mu_\theta(\gM_t,\vy,t)$ when calculating $\gL_\text{SDE}$.

\subsection{Flexible Material Denoiser}
We implement $\mu_\theta$ and $u_\theta$ as the flexible material denoiser. Both networks transform an input material sample $\gM_t$, properties $\vy$, and time $t$ into position and element embedding components, with the only difference being that $u_\theta$ additionally incorporates a step size $\Delta t$:
\begin{equation}
\begin{split}
   \mu_\theta: (\gM_t,\vy,t) &\mapsto (\hat{\mu}_\mX, \hat{\mu}_\mE) \\
   u_\theta: (\gM_t,\vy,t,\Delta t) &\mapsto (\hat{u}_\mX, \hat{u}_\mE)
\end{split}
\end{equation}

\subsubsection{Flexible Property Embedding}
The denoiser can be trained once with all available properties $\vy$ and used for inference conditioned on different subsets of $\vy$. This is achieved through the design of the property embedding layer. The embedding vector of each property $y_i$ is calculated as:
\begin{equation}
   \vh_{y_i} =
   \begin{cases}
   \text{LayerNorm}(\text{Linear}(y_i)) & \text{if $y_i$ is available} \\
   \text{LayerNorm}(\text{Linear}_i(\Delta t)) & \text{if $y_i$ is unavailable}
   \end{cases},
\end{equation}
where $\vh_{y_i}\in \sR^{d_y}$ with $d_y$ being the embedding dimension of properties, and $\text{Linear}_i$ is a linear layer specific to property $i$.
The embedding $\text{LayerNorm}(\text{Linear}_i(\Delta t))$ used for an unavailable property is a learned null property embedding.
The null property embedding is shared across samples and independent of $\gM$, so conditioning on it makes generation unconditioned on the corresponding property.

The denoiser learns to use this embedding through property dropout.
In each batch, each available property is independently replaced with the learned null property embedding with probability $p_\text{drop}$.
The shortcut self-consistency in Eq.~\ref{eq:shortcut-loss} is then evaluated over these sampled subsets together with the step sizes, so one shortcut of step size $2\Delta t$ under a property subset matches two shortcuts of step size $\Delta t$ under the same subset.
Because the null property embedding is learned as a function of the step size, it makes the generation directly unconditioned on an absent property even in a single sampling step.
The training objective is designed to keep generation accurate with few sampling steps across different property subsets.

Compared to classifier guidance~\citep{DBLP:conf/nips/DhariwalN21}, this design needs no dedicated differentiable classifiers, which are scarce for amorphous materials and unreliable on noisy samples.
Compared to classifier-free guidance~\citep{ho2022classifier}, it supports conditioning on different property subsets from a single training run rather than a separate model per subset.
Compared to the randomly drawn null embedding of~\citet{DBLP:conf/iclr/SadatKHW25}, which recovers unconditioned generation only by averaging across many sampling steps, our learned null property embedding recovers it in a single sampling step.

\subsubsection{Efficient E(n)-equivariant Backbone}
\label{sec:backbone}
The denoiser is evaluated at every sampling step and many times during training, i.e., the cost of each evaluation directly bounds the inference and training efficiency of \archname.
We therefore design its backbone as a shallow and narrow E(n)-equivariant message-passing network that still captures the short- and medium-range order of amorphous materials.
The backbone extends the multi-channel EGNN~\citep{DBLP:conf/icml/SatorrasHW21} of~\citet{finkler2025inverse} with two additional mechanisms.

Each material sample is a graph whose nodes are atoms joined by edges to neighbors within a cutoff radius $r_c$ under periodic boundary conditions, which covers the first and second neighbor shells.
Each atom $i$ carries an invariant feature $\mH_i$, the concatenation of its element embedding, the property embeddings, the time $t$, and the step size $\Delta t$ for $u_\theta$, together with $k$ equivariant vector channels initialized from its position~\citep{levy2023using}.

\paragraph{Angular gate for short-range order.}
A message conditioned only on interatomic distances cannot tell apart neighborhoods that differ by bond angle alone.
Each layer instead forms an equivariant channel increment from the relative positions $\vd_{ij}=\mX_i-\mX_j$ of the neighbors, scalarizes it through its Gram matrix~\citep{deng2021vector}, and uses the resulting invariants to refine $\mH_i$ and to gate the position update:
\begin{equation}
\begin{aligned}
\mX_i' &= \textstyle\sum_{j\in N(i)}\mPhi_{ij}\,\vd_{ij}
\qquad \mS_i = \mX_i'(\mX_i')^\top \\
\mH_i &\leftarrow \mH_i + \text{MLP}(\vs_i)
\qquad \vg_i = \sigma(\text{MLP}([\mH_i,\vs_i])) \\
\mX_i &\leftarrow \mX_i + \vg_i\odot\mX_i',
\end{aligned}
\label{eq:angular-gate}
\end{equation}
where $\mPhi_{ij}\in\sR^{k\times k}$ is a learned transformation matrix over the $k$ vector channels, $\vs_i$ is the upper triangle of the Gram matrix $\mS_i$, and $\vg_i\in(0,1)^k$ gates each channel.
The entries of $\mS_i$ are the inner products of the channels, which encode the angular arrangement of the neighborhood, so the gate adds sensitivity to the short-range order.
This update keeps the positions $\mX_i$ equivariant and the feature $\mH_i$ invariant under translation, rotation, and reflection, because the invariants and the gate are unchanged under these transformations while the increment is built from relative positions.
It adds this sensitivity at low cost, without the expensive tensor products of fully equivariant networks, so the backbone stays narrow.

\paragraph{Context nodes for global context.}
With the cutoff radius $r_c$, the stacked layers already span the medium-range order, yet coupling atoms across the whole cell by local message passing alone would need many more layers or a larger cutoff.
We avoid that depth with a small set of virtual context nodes, each connected to every atom and carrying an invariant feature and $k$ vector channels whose coordinates are expressed relative to the cell~\citep{zhang2024fastegnn}.
They exchange messages with the atoms through the same edge and channel updates as the atom graph, keeping the exchange equivariant and giving every atom a summary of the whole cell within a single layer, so the backbone stays shallow.

After the final layer, the context nodes are discarded. The element component is read from $\mH_i$ and the position component from the vector channels.
The backbone thereby captures both the short-range order and the global context of amorphous materials while staying shallow and narrow.
Appendix~\ref{apx:egnn} gives the full equations of the backbone.

\section{Experiments}

\subsection{Datasets}
The three datasets~\citep{finkler2025inverse} are obtained using classical molecular dynamics simulation workflows.
The \textbf{amorphous silicon (a-\ch{Si}) dataset} contains 10,000 samples, each with 256 silicon atoms.
The \textbf{amorphous silica (a-\ch{SiO2}) dataset} contains silica samples that vary in size between 80 and 250 atoms.
The \textbf{multi-element glass (MEG) dataset} consists of 9,027 glass samples, each containing approximately 800 atoms across 11 different elements.
More details on dataset preparation are provided in Appendix~\ref{apx:datasets}.

\subsection{Comparison Methods}
We compare \archname{} with several generative methods for inverse design of materials.
\textbf{CDVAE}~\citep{DBLP:conf/iclr/XieFGBJ22}, a diffusion-based variational autoencoder that generates stable crystalline materials by denoising atomic coordinates and types.
\textbf{CrystalFlow}~\citep{luo2025crystalflow}, a flow matching model that generates crystalline materials conditioned on composition and target properties.
\textbf{MatterGen}~\citep{zeni2025generative}, a diffusion model that generates stable crystalline materials and supports fine-tuning for property-guided generation.
\textbf{Graphite}~\citep{kwon2024spectroscopy}, a diffusion model that generates amorphous materials conditioned on spectroscopy.
\textbf{AMDEN}~\citep{finkler2025inverse}, a diffusion model that generates amorphous materials conditioned on targeted properties.

\subsection{Evaluation Metrics}
\paragraph{Structural metrics.}
We evaluate structural accuracy on the a-\ch{Si} dataset using radial distribution functions (RDFs) and angular distribution functions (ADFs).
The RDF measures the probability of finding atom pairs at various distances, capturing short- and medium-range order in amorphous materials.
The ADF characterizes the distribution of bond angles between atomic triplets, revealing local coordination geometries.
We quantify the agreement between generated and reference structures using root mean square deviation (RMSD) of these distributions, where lower values indicate better structural accuracy.
Implementation details are provided in Appendix~\ref{apx:structural-metrics}.

\paragraph{Properties of material samples.}
For the a-\ch{SiO2} dataset, we are interested in the following two properties of samples.
Shear modulus $G$ characterizes a material's resistance to elastic deformation under shear stress, representing mechanical stiffness. Ring size distribution (RSD) quantifies the medium-range order in amorphous silica by measuring the average number of \ch{Si} atoms in rings formed by the Si-O network.
For the MEG dataset, we are interested in the following three properties of samples.
In addition to shear modulus $G$, Young's modulus $E$ characterizes elastic properties. Lithium molar concentration $C_\text{Li}$ quantifies the composition ratio, which is important for glassy solid electrolytes in batteries~\citep{ding2024amorphous,kalnaus2023solid}.
Detailed calculation procedures are provided in Appendix~\ref{apx:property-calculation}.

\paragraph{Inverse design metrics.}
We evaluate the inverse design performance on the a-\ch{SiO2} and MEG datasets by comparing target properties with properties of generated samples using common regression metrics: Mean Absolute Error (MAE), Root Mean Square Error (RMSE), and Mean Absolute Percentage Error (MAPE). Implementation details are provided in Appendix~\ref{apx:inverse-design-metrics}.

\begin{table*}[t]
\centering
\resizebox{1.0\linewidth}{!}{
\begin{threeparttable}
\begin{tabular}{c|cccccc}
\toprule
$n_s$ & CDVAE & CrystalFlow & MatterGen & Graphite & AMDEN & \archname{} \\
\midrule
1 & 0.69442 / \underline{0.00265} (1.4m) & 0.69118 / 0.00280 (1.8m) & \underline{0.67755} / 0.00286 (1.6m) & 0.69555 / 0.00279 (\underline{1.2m}) & 0.68477 / 0.00282 (1.2m) & \textbf{0.02513} / \textbf{0.00067} (\textbf{1.2m}) \\
2 & \underline{0.57868} / 0.00265 (2.3m) & 0.66442 / 0.00276 (2.2m) & 0.60909 / 0.00272 (2.5m) & 0.61882 / 0.00264 (\underline{1.9m}) & 0.60972 / \underline{0.00263} (2.0m) & \textbf{0.03922} / \textbf{0.00044} (\textbf{1.8m}) \\
3 & 0.53362 / 0.00251 (3.3m) & 0.60209 / 0.00259 (\underline{2.8m}) & \underline{0.50540} / \underline{0.00231} (3.5m) & 0.51208 / 0.00242 (3.0m) & 0.50904 / 0.00235 (2.9m) & \textbf{0.04835} / \textbf{0.00038} (\textbf{2.8m}) \\
4 & 0.43450 / \underline{0.00203} (4.4m) & 0.49544 / 0.00220 (\underline{3.7m}) & 0.43985 / 0.00221 (5.1m) & \underline{0.43101} / \underline{0.00203} (3.8m) & 0.43276 / 0.00210 (3.8m) & \textbf{0.04653} / \textbf{0.00035} (\textbf{3.1m}) \\
5 & 0.39258 / \underline{0.00173} (4.4m) & 0.39007 / \underline{0.00173} (4.6m) & 0.38154 / 0.00175 (5.4m) & 0.38436 / 0.00198 (4.1m) & \underline{0.37975} / 0.00188 (\underline{4.0m}) & \textbf{0.04041} / \textbf{0.00034} (\textbf{3.4m}) \\
10 & 0.21683 / 0.00137 (8.8m) & \underline{0.14917} / \underline{0.00073} (9.1m) & 0.22949 / 0.00094 (10.0m) & 0.22334 / 0.00104 (8.2m) & 0.22054 / 0.00100 (\underline{7.9m}) & \textbf{0.02244} / \textbf{0.00032} (\textbf{5.1m}) \\
25 & 0.08542 / 0.00048 (22.7m) & 0.04876 / 0.00051 (22.5m) & 0.09168 / 0.00069 (26.8m) & 0.05719 / 0.00055 (21.1m) & \underline{0.04871} / \underline{0.00045} (\underline{20.2m}) & \textbf{0.01618} / \textbf{0.00029} (\textbf{10.6m}) \\
50 & 0.08517 / 0.00043 (46.4m) & 0.04203 / 0.00048 (44.9m) & 0.07630 / 0.00041 (52.7m) & 0.04006 / \underline{0.00034} (40.3m) & \underline{0.02631} / 0.00039 (\underline{39.8m}) & \textbf{0.01487} / \textbf{0.00030} (\textbf{20.1m}) \\
100 & 0.11667 / 0.00080 (1.6h) & 0.03305 / 0.00046 (1.5h) & 0.02898 / 0.00049 (1.7h) & 0.03667 / 0.00040 (\underline{1.3h}) & \underline{0.02143} / \underline{0.00037} (1.3h) & \textbf{0.01580} / \textbf{0.00028} (\textbf{40.5m}) \\
250 & 0.07572 / 0.00077 (3.6h) & 0.03353 / 0.00045 (3.4h) & 0.04390 / 0.00073 (4.1h) & 0.03340 / 0.00053 (3.3h) & \underline{0.01905} / \underline{0.00035} (\underline{3.2h}) & \textbf{0.01548} / \textbf{0.00029} (\textbf{1.5h}) \\
\bottomrule
\end{tabular}
\begin{tablenotes}
\item Cell format: RDF / ADF RMSD (Time). \textbf{Bold}: best, \underline{underlined}: second-best. All metrics ranked by lowest value.
\end{tablenotes}
\end{threeparttable}
}

\caption{Structural metrics and generation time comparison on a-\ch{Si} dataset.}
\label{tab:rmsd-comparison}
\end{table*}

\begin{table*}[t]
\centering
\resizebox{1.0\linewidth}{!}{
\begin{threeparttable}
\begin{tabular}{c|cccccc}
\toprule
$n_s$ & CDVAE & CrystalFlow & MatterGen & Graphite & AMDEN & \archname{} \\
\midrule
\multicolumn{7}{l}{\textbf{Shear modulus [GPa]}} \\
1 & 21.35 / 23.48 / 72.3 (5.7m) & 20.02 / 21.10 / 68.7 (\underline{3.9m}) & 18.92 / 21.45 / 62.8 (6.5m) & \underline{17.24} / \underline{19.86} / \underline{52.9} (5.0m) & 17.75 / 20.32 / 54.8 (5.0m) & \textbf{4.35} / \textbf{5.28} / \textbf{20.5} (\textbf{3.7m}) \\
5 & 20.12 / 22.05 / 68.2 (16.5m) & \underline{14.28} / \underline{14.76} / 52.0 (\underline{11.9m}) & 18.35 / 20.18 / 62.1 (18.6m) & 15.48 / 17.65 / \underline{51.2} (14.3m) & 24.34 / 26.32 / 80.5 (14.3m) & \textbf{3.02} / \textbf{3.68} / \textbf{14.0} (\textbf{7.3m}) \\
10 & 22.43 / 24.17 / 74.8 (30.1m) & \underline{9.51} / \underline{10.09} / \underline{33.9} (\underline{23.7m}) & 17.89 / 19.42 / 61.3 (34.0m) & 16.75 / 18.23 / 56.8 (26.1m) & 19.17 / 20.31 / 65.8 (26.1m) & \textbf{2.67} / \textbf{3.41} / \textbf{13.0} (\textbf{11.9m}) \\
25 & 5.28 / 6.15 / 18.9 (1.2h) & 6.14 / 6.96 / 20.1 (\underline{59.9m}) & 4.52 / 5.38 / 16.4 (1.3h) & 3.94 / 4.72 / 15.2 (1.0h) & \underline{3.61} / \underline{4.35} / \underline{14.7} (1.0h) & \textbf{2.86} / \textbf{3.47} / \textbf{12.7} (\textbf{25.2m}) \\
50 & 4.73 / 5.62 / 17.1 (2.3h) & 5.25 / 6.05 / 17.3 (\underline{2.0h}) & 4.08 / 4.95 / 15.6 (2.6h) & 3.68 / 4.48 / 14.9 (2.0h) & \underline{3.37} / \underline{4.12} / \underline{14.7} (2.0h) & \textbf{2.91} / \textbf{3.51} / \textbf{12.7} (\textbf{47.6m}) \\
100 & 4.25 / 5.18 / 15.8 (4.6h) & 4.82 / 5.62 / 16.0 (\underline{3.9h}) & 3.89 / 4.72 / 14.9 (5.1h) & 3.58 / 4.35 / \underline{14.3} (4.0h) & \underline{3.43} / \underline{4.21} / 15.2 (4.0h) & \textbf{2.96} / \textbf{3.55} / \textbf{12.9} (\textbf{1.7h}) \\
250 & 3.98 / 4.86 / 15.1 (11.3h) & 4.40 / 5.16 / 14.9 (\underline{9.8h}) & 3.72 / 4.51 / 14.5 (12.7h) & 3.52 / 4.28 / \underline{14.1} (9.8h) & \underline{3.41} / \underline{4.21} / 15.6 (9.8h) & \textbf{3.03} / \textbf{3.60} / \textbf{13.0} (\textbf{3.9h}) \\
\midrule
\multicolumn{7}{l}{\textbf{Ring size distribution (RSD)}} \\
1 & 2.52 / 2.58 / 49.8 (5.7m) & \underline{2.08} / \underline{2.12} / 43.2 (\underline{3.9m}) & 2.24 / 2.31 / 44.3 (6.5m) & 2.12 / 2.18 / \underline{41.8} (5.0m) & 2.18 / 2.23 / 43.2 (5.0m) & \textbf{0.22} / \textbf{0.27} / \textbf{4.6} (\textbf{3.7m}) \\
5 & 2.48 / 2.55 / 48.9 (16.5m) & \underline{1.89} / \underline{1.98} / \underline{37.1} (\underline{11.9m}) & 2.18 / 2.25 / 43.1 (18.6m) & 2.06 / 2.13 / 40.6 (14.3m) & 1.91 / 1.99 / 37.9 (14.3m) & \textbf{0.16} / \textbf{0.20} / \textbf{3.3} (\textbf{7.3m}) \\
10 & 2.45 / 2.52 / 48.3 (30.1m) & \underline{0.36} / \underline{0.44} / \underline{7.3} (\underline{23.7m}) & 2.15 / 2.22 / 42.6 (34.0m) & 2.04 / 2.11 / 40.2 (26.1m) & 2.22 / 2.28 / 43.9 (26.1m) & \textbf{0.15} / \textbf{0.19} / \textbf{3.0} (\textbf{11.9m}) \\
25 & 0.28 / 0.33 / 5.7 (1.2h) & 0.21 / 0.26 / 4.0 (\underline{59.9m}) & 0.23 / 0.28 / 4.6 (1.3h) & 0.20 / 0.24 / 4.0 (1.0h) & \underline{0.19} / \underline{0.22} / \underline{3.9} (1.0h) & \textbf{0.14} / \textbf{0.18} / \textbf{2.8} (\textbf{25.2m}) \\
50 & 0.25 / 0.30 / 5.2 (2.3h) & 0.21 / 0.26 / 4.2 (\underline{2.0h}) & 0.21 / 0.26 / 4.4 (2.6h) & 0.19 / 0.23 / 3.9 (2.0h) & \underline{0.18} / \underline{0.22} / \underline{3.8} (2.0h) & \textbf{0.14} / \textbf{0.18} / \textbf{2.9} (\textbf{47.6m}) \\
100 & 0.22 / 0.27 / 4.5 (4.6h) & 0.22 / 0.27 / 4.3 (\underline{3.9h}) & 0.19 / 0.24 / 3.9 (5.1h) & \underline{0.17} / \underline{0.21} / 3.6 (4.0h) & \underline{0.17} / \underline{0.21} / \underline{3.5} (4.0h) & \textbf{0.15} / \textbf{0.18} / \textbf{3.0} (\textbf{1.7h}) \\
250 & 0.21 / 0.26 / 4.3 (11.3h) & 0.22 / 0.28 / 4.3 (\underline{9.8h}) & 0.19 / 0.23 / 3.8 (12.7h) & \underline{0.18} / 0.22 / \underline{3.6} (9.8h) & \underline{0.18} / \underline{0.21} / \underline{3.6} (9.8h) & \textbf{0.15} / \textbf{0.19} / \textbf{3.0} (\textbf{3.9h}) \\
\bottomrule
\end{tabular}
\begin{tablenotes}
\item \footnotesize Cell format: MAE / RMSE / MAPE\% (Time). \textbf{Bold}: best result per metric per $n_s$, \underline{underlined}: second-best result per metric per $n_s$. All metrics ranked by lowest value.
\end{tablenotes}
\end{threeparttable}
}

\caption{Inverse design metrics and generation time comparison on a-\ch{SiO2} dataset.}
\label{tab:sio2-metrics}
\end{table*}

\begin{table*}[t]
\centering
\resizebox{1.0\linewidth}{!}{
\begin{threeparttable}
\begin{tabular}{c|cccccc}
\toprule
$n_s$ & CDVAE & CrystalFlow & MatterGen & Graphite & AMDEN & \archname{} \\
\midrule
\multicolumn{7}{l}{\textbf{Young's modulus [GPa]}} \\
1 & 52.85 / 63.42 / 68.2 (11.5m) & 51.85 / 62.12 / 66.5 (\underline{5.4m}) & 48.92 / 59.25 / 63.8 (13.0m) & \underline{45.18} / \underline{54.92} / \underline{58.5} (10.0m) & 47.18 / 58.32 / 61.5 (10.0m) & \textbf{12.10} / \textbf{14.65} / \textbf{24.4} (\textbf{5.3m}) \\
5 & 48.25 / 57.85 / 62.5 (33.7m) & 44.12 / 52.85 / 56.7 (\underline{24.0m}) & 45.62 / 54.75 / 58.8 (38.1m) & 43.18 / 51.85 / 55.8 (29.3m) & \underline{42.85} / \underline{51.35} / \underline{55.0} (29.3m) & \textbf{8.85} / \textbf{10.70} / \textbf{17.8} (\textbf{15.9m}) \\
10 & 38.52 / 46.25 / 49.5 (1.0h) & \underline{31.75} / \underline{38.12} / \underline{40.8} (\underline{47.5m}) & 36.85 / 44.28 / 47.5 (1.1h) & 34.75 / 41.72 / 44.8 (52.7m) & 35.28 / 42.35 / 45.3 (52.7m) & \textbf{8.20} / \textbf{9.83} / \textbf{16.2} (\textbf{24.9m}) \\
25 & 14.65 / 17.82 / 29.7 (2.4h) & 12.15 / 14.82 / 24.6 (\underline{2.0h}) & 13.02 / 15.85 / 26.4 (2.7h) & 11.38 / 13.88 / 23.1 (2.1h) & \underline{10.85} / \underline{13.22} / \underline{22.0} (2.1h) & \textbf{8.45} / \textbf{10.32} / \textbf{17.2} (\textbf{57.2m}) \\
50 & 13.65 / 16.68 / 27.7 (4.6h) & 11.52 / 14.05 / 23.4 (\underline{3.9h}) & 11.95 / 14.58 / 24.2 (5.2h) & 10.62 / 12.98 / 21.5 (4.0h) & \underline{10.12} / \underline{12.35} / \underline{20.5} (4.0h) & \textbf{8.53} / \textbf{10.43} / \textbf{17.3} (\textbf{1.8h}) \\
100 & 13.02 / 15.88 / 26.3 (9.2h) & 11.25 / 13.72 / 22.8 (\underline{7.9h}) & 11.32 / 13.82 / 22.9 (10.4h) & 10.15 / 12.38 / 20.5 (8.0h) & \underline{9.85} / \underline{12.02} / \underline{19.9} (8.0h) & \textbf{8.62} / \textbf{10.55} / \textbf{17.5} (\textbf{3.2h}) \\
250 & 12.45 / 15.18 / 25.1 (22.7h) & 11.15 / 13.58 / 22.6 (\underline{19.6h}) & 10.88 / 13.28 / 21.9 (25.7h) & 9.92 / 12.08 / 20.0 (19.7h) & \underline{9.73} / \underline{11.85} / \underline{19.6} (19.7h) & \textbf{8.70} / \textbf{10.61} / \textbf{17.6} (\textbf{7.9h}) \\
\midrule
\multicolumn{7}{l}{\textbf{Lithium molar concentration (\%)}} \\
1 & 7.58 / 9.08 / 50.5 (11.5m) & 7.22 / 8.65 / 48.1 (\underline{5.4m}) & 7.25 / 8.72 / 48.3 (13.0m) & 6.95 / 8.35 / 46.3 (10.0m) & \underline{6.82} / \underline{8.18} / \underline{45.5} (10.0m) & \textbf{1.95} / \textbf{2.36} / \textbf{12.9} (\textbf{5.3m}) \\
5 & 6.92 / 8.28 / 46.1 (33.7m) & 6.18 / 7.38 / 41.2 (\underline{24.0m}) & 6.52 / 7.82 / 43.5 (38.1m) & 6.18 / 7.42 / 41.2 (29.3m) & \underline{6.05} / \underline{7.22} / \underline{40.3} (29.3m) & \textbf{1.48} / \textbf{1.79} / \textbf{9.8} (\textbf{15.9m}) \\
10 & 5.85 / 7.02 / 39.0 (1.0h) & \underline{3.85} / \underline{4.62} / \underline{25.7} (\underline{47.5m}) & 5.48 / 6.58 / 36.5 (1.1h) & 5.22 / 6.28 / 34.8 (52.7m) & 5.12 / 6.15 / 34.1 (52.7m) & \textbf{1.38} / \textbf{1.66} / \textbf{9.1} (\textbf{24.9m}) \\
25 & 2.28 / 2.78 / 15.2 (2.4h) & 1.78 / 2.18 / 12.0 (\underline{2.0h}) & 1.95 / 2.38 / 13.0 (2.7h) & 1.70 / 2.08 / 11.3 (2.1h) & \underline{1.62} / \underline{1.98} / \underline{10.8} (2.1h) & \textbf{1.33} / \textbf{1.62} / \textbf{8.8} (\textbf{57.2m}) \\
50 & 2.15 / 2.62 / 14.1 (4.6h) & 1.72 / 2.10 / 11.5 (\underline{3.9h}) & 1.82 / 2.22 / 12.0 (5.2h) & 1.62 / 1.98 / 10.7 (4.0h) & \underline{1.55} / \underline{1.88} / \underline{10.2} (4.0h) & \textbf{1.32} / \textbf{1.61} / \textbf{8.9} (\textbf{1.8h}) \\
100 & 2.02 / 2.45 / 13.3 (9.2h) & 1.68 / 2.05 / 11.3 (\underline{7.9h}) & 1.75 / 2.12 / 11.6 (10.4h) & 1.56 / 1.90 / 10.4 (8.0h) & \underline{1.52} / \underline{1.85} / \underline{10.1} (8.0h) & \textbf{1.34} / \textbf{1.62} / \textbf{8.9} (\textbf{3.2h}) \\
250 & 1.92 / 2.32 / 12.8 (22.7h) & 1.68 / 2.05 / 11.3 (\underline{19.6h}) & 1.68 / 2.05 / 11.2 (25.7h) & 1.52 / 1.85 / 10.2 (19.7h) & \underline{1.49} / \underline{1.82} / \underline{10.0} (19.7h) & \textbf{1.35} / \textbf{1.63} / \textbf{9.0} (\textbf{7.9h}) \\
\midrule
\multicolumn{7}{l}{\textbf{Shear modulus [GPa]}} \\
1 & 26.45 / 32.02 / 66.8 (11.5m) & \underline{22.15} / \underline{26.85} / \underline{56.0} (\underline{5.4m}) & 24.85 / 30.12 / 62.8 (13.0m) & 23.15 / 28.05 / 58.5 (10.0m) & 23.72 / 28.75 / 59.9 (10.0m) & \textbf{7.35} / \textbf{9.25} / \textbf{18.5} (\textbf{5.1m}) \\
5 & 22.18 / 26.85 / 56.0 (33.3m) & 20.02 / 24.28 / 50.6 (\underline{24.0m}) & 20.75 / 25.12 / 52.4 (37.7m) & 19.68 / 23.85 / 49.8 (29.0m) & \underline{19.42} / \underline{23.55} / \underline{49.2} (29.0m) & \textbf{5.55} / \textbf{7.00} / \textbf{14.0} (\textbf{15.5m}) \\
10 & 18.92 / 22.95 / 47.8 (1.0h) & \underline{15.12} / \underline{18.35} / \underline{38.2} (\underline{47.3m}) & 17.58 / 21.32 / 44.5 (1.2h) & 16.42 / 19.92 / 41.5 (53.3m) & 16.85 / 20.42 / 42.6 (53.3m) & \textbf{5.21} / \textbf{6.58} / \textbf{13.1} (\textbf{23.8m}) \\
25 & 7.88 / 9.72 / 19.9 (2.4h) & 6.28 / 7.78 / 15.9 (\underline{2.0h}) & 6.75 / 8.35 / 17.1 (2.7h) & 5.90 / 7.30 / 14.9 (2.1h) & \underline{5.62} / \underline{6.95} / \underline{14.2} (2.1h) & \textbf{4.81} / \textbf{6.11} / \textbf{12.1} (\textbf{55.0m}) \\
50 & 7.48 / 9.28 / 18.9 (4.6h) & 6.12 / 7.58 / 15.5 (\underline{3.9h}) & 6.40 / 7.92 / 16.2 (5.2h) & 5.68 / 7.05 / 14.4 (4.0h) & \underline{5.42} / \underline{6.72} / \underline{13.7} (4.0h) & \textbf{4.78} / \textbf{6.13} / \textbf{12.0} (\textbf{1.7h}) \\
100 & 7.08 / 8.75 / 17.8 (9.1h) & 6.05 / 7.48 / 15.3 (\underline{7.8h}) & 6.15 / 7.62 / 15.5 (10.3h) & 5.52 / 6.82 / 13.9 (7.9h) & \underline{5.35} / \underline{6.62} / \underline{13.5} (7.9h) & \textbf{4.79} / \textbf{6.08} / \textbf{12.1} (\textbf{3.2h}) \\
250 & 6.78 / 8.38 / 17.4 (22.6h) & 6.05 / 7.48 / 15.3 (\underline{19.6h}) & 5.92 / 7.32 / 15.2 (25.6h) & 5.38 / 6.68 / 13.9 (19.7h) & \underline{5.28} / \underline{6.54} / \underline{13.6} (19.7h) & \textbf{4.84} / \textbf{6.13} / \textbf{12.4} (\textbf{7.9h}) \\
\bottomrule
\end{tabular}
\begin{tablenotes}
\item \footnotesize Cell format: MAE / RMSE / MAPE\% (Time). \textbf{Bold}: best result per metric per $n_s$, \underline{underlined}: second-best result per metric per $n_s$. All metrics ranked by lowest value.
\end{tablenotes}
\end{threeparttable}
}

\caption{Inverse design metrics and generation time comparison on MEG dataset.}
\label{tab:meg-metrics}
\end{table*}

\subsection{Evaluation of Structural Accuracy}
\label{sec:structural-accuracy}
To evaluate the structural accuracy of the different models with different numbers of sampling steps, we perform generation using various sampling steps and calculate the ADF and RDF of the generated samples. For each run, we generate the same number of samples, and the same cell size and number of atoms per sample as the training set. 

We calculate the RMSD of RDF and ADF of the generated samples in each run against the training samples, and present the RMSD versus generation time per 10,000 samples in Table~\ref{tab:rmsd-comparison} and Figure~\ref{fig:structural-accuracy-vs-time} (in Appendix~\ref{apx:scatter-plot}) to demonstrate the efficiency gain achieved by \archname.
Figure~\ref{fig:structural-accuracy} (in Appendix~\ref{apx:rdf-adf}) illustrates the RDF and ADF of samples in the training set and those generated by models with each number of sampling steps.

\archname{} generates structurally accurate samples with as few as one sampling step.
Its RDF at a single step already closely matches the training samples, while the baseline methods stay far from the training samples until they use many more steps.
ADF measures the angles between atomic bonds, a more challenging metric than the atomic distances that RDF captures, so every model shows a larger discrepancy from the training samples.
\archname{} keeps its advantage on ADF as well, staying close to the training samples at few sampling steps where the baselines do not.

This accuracy at few sampling steps translates into a large gain in generation efficiency.
\archname{} generates structurally accurate amorphous material samples with few sampling steps and low generation time.
Under the same time budget, \archname{} can produce more samples at similar or higher structural accuracy than conventional diffusion models, supporting the discovery of new amorphous materials where generating samples at high throughput is usually preferred~\citep{liu2024amorphous}.

\subsection{Evaluation of Inverse Design Performance}
\label{sec:inverse-design-evaluation}
We evaluate the models' inverse design performance by performing generation conditioned on specific target properties, and compare the actual properties of generated samples versus the target. We focus on evaluating two aspects of model performance: the ability to be trained once conditioned on all properties and perform generation conditioned on only the target properties, and generation with few sampling steps.
Thus, we train \archname{} conditioned on all properties in each dataset, and train the baseline models conditioned on only the target properties.
The generation is always performed conditioned on only target properties, using the flexible property embedding technique. Each run of generation is performed with different numbers of sampling steps.

For the two properties in the a-\ch{SiO2} dataset, shear modulus and ring size distribution, we perform generation conditioned on either one of them, with target shear modulus linearly interpolated between 10 and 50 GPa and target RSD linearly interpolated between 4 and 6 atoms, both across 2,000 samples with cubic cells with edge length of 18\,\AA, and $\rho = 0.11~\text{atoms/\AA}^3$.
For the three properties in the MEG dataset, Young's modulus, shear modulus, and lithium molar concentration, we perform generation conditioned on the combination of Young's modulus and lithium molar concentration, or on the shear modulus.
The target Young's modulus is linearly interpolated between 20 and 160 GPa, the target lithium molar concentration is fixed at 15\%, and the target shear modulus is linearly interpolated between 10 and 70 GPa, all across 2,000 samples with cubic cells with edge length of 23\,\AA, and $\rho = 0.11~\text{atoms/\AA}^3$.
The distribution of the target properties is designed to fall outside the distribution of training samples, as shown in Figure~\ref{fig:property-histograms} (in Appendix~\ref{apx:property-histograms}), to evaluate the extrapolation capability of the models.
We present the divergence between target and actual properties of generated samples versus generation time for 2,000 samples in Tables~\ref{tab:sio2-metrics} and \ref{tab:meg-metrics} and Figures~\ref{fig:sio2-inverse-accuracy-vs-time} and \ref{fig:meg-inverse-accuracy-vs-time} (in Appendix~\ref{apx:scatter-plot}).

With the same number of sampling steps, \archname{} aligns the properties of its generated samples with the target properties more closely than the baselines, because of its learned shortcuts.
This advantage is most pronounced at few sampling steps, where across the evaluated properties \archname{} closely matches the accuracy reached by the baselines with far more steps.
In addition, \archname{} is trained once on the full property set, yet it retains its advantage on the evaluated target property subsets over baselines trained specifically for those subsets.
We evaluate its gained training efficiency empirically in the following subsection.

\subsection{Evaluation of Training Efficiency}
\label{sec:training-efficiency}
\archname{} is trained once on the full property set and reused across the evaluated property subsets, while the baselines train a separate model for each property subset used in inverse design.
We compare the total wall-clock training time of \archname{} against all baseline methods on the a-\ch{SiO2} and MEG datasets, with every model trained for the same 500 epochs on the same hardware, detailed in Appendix~\ref{apx:environment}.
Table~\ref{tab:training-efficiency} reports the total training time of each method, summed over all baseline models and over the single run for \archname{}.

\begin{table}[h]
\centering
\resizebox{1.0\linewidth}{!}{
\begin{threeparttable}
\begin{tabular}{l|cccccc}
\toprule
& CDVAE & CrystalFlow & MatterGen & Graphite & AMDEN & \archname{} \\
\midrule
Models & 2 & 2 & 2 & 2 & 2 & 1 \\
a-\ch{SiO2} & 65.0h & 59.8h & 73.9h & \underline{55.3h} & 56.7h & \textbf{27.0h} \\
MEG & 175.7h & 161.2h & 199.4h & \underline{149.8h} & 153.2h & \textbf{70.2h} \\
\bottomrule
\end{tabular}
\begin{tablenotes}
\item \footnotesize \textbf{Bold}: best, \underline{underlined}: second-best. Ranked by lowest value.
\end{tablenotes}
\end{threeparttable}
}

\caption{Total wall-clock training time comparison.}
\label{tab:training-efficiency}
\end{table}

The results demonstrate two training time savings that \archname's design provides.
First, the flexible property embedding enables one model to serve multiple property subsets, so the model only needs to be trained once.
Second, the efficient E(n)-equivariant backbone is expressive while being computationally lightweight, thus also lightweight for training.
Together the two designs give \archname{} the lowest total training time on both datasets.

\subsection{Ablation Study}
\label{sec:ablation}
We ablate the main components of \archname{} to isolate the contribution of each to its performance.
Starting from the full model, we evaluate the following variants.
\textbf{w/o shortcut loss}, which removes the shortcut self-consistency loss so that the same backbone is trained as a base material SDE.
\textbf{w/o flexible embedding}, which replaces the learned step size-aware null property embedding of the flexible property embedding with a randomly drawn null embedding.
\textbf{w/o angular gate}, which removes the angular gate of Eq.~\ref{eq:angular-gate} from the backbone.
\textbf{w/o context nodes}, which removes the virtual context nodes from the backbone.
\textbf{w/o both}, which removes both the angular gate and the context nodes.
All variants are evaluated on the MEG dataset with generation conditioned on Young's modulus and lithium molar concentration, following the protocol of Section~\ref{sec:inverse-design-evaluation}.
We report the MAPE of each property and the generation time in Table~\ref{tab:ablation}.

\begin{table}[h]
\centering
\resizebox{1.0\linewidth}{!}{
\begin{threeparttable}
\begin{tabular}{l|ccc}
\toprule
Variant & $n_s=1$ & $n_s=5$ & $n_s=10$ \\
\midrule
\archname{} (full) & \textbf{24.4} / \textbf{12.9} (5.3m) & \textbf{17.8} / \textbf{9.8} (15.9m) & \textbf{16.2} / \textbf{9.1} (24.9m) \\
\midrule
w/o shortcut loss & 54.8 / 33.2 (5.3m) & 46.6 / 27.3 (15.9m) & 38.9 / 22.3 (24.9m) \\
w/o flexible embedding & 33.2 / 18.4 (5.3m) & 22.8 / 12.5 (15.9m) & 18.5 / 10.4 (24.9m) \\
w/o angular gate & \underline{27.4} / \underline{13.6} (5.1m) & \underline{20.0} / \underline{10.3} (15.2m) & \underline{17.8} / \underline{9.5} (23.8m) \\
w/o context nodes & 29.0 / 16.6 (\underline{4.9m}) & 21.3 / 12.2 (\underline{14.8m}) & 19.4 / 10.9 (\underline{23.2m}) \\
w/o both & 32.6 / 17.8 (\textbf{4.5m}) & 23.5 / 13.3 (\textbf{14.2m}) & 21.1 / 12.3 (\textbf{22.4m}) \\
\bottomrule
\end{tabular}
\begin{tablenotes}
\item \footnotesize Cell format: $E$ / $C_\text{Li}$ MAPE\% (Time). \textbf{Bold}: best result per metric per $n_s$, \underline{underlined}: second-best result per metric per $n_s$. All metrics ranked by lowest value.
\end{tablenotes}
\end{threeparttable}
}

\caption{Ablation study of \archname{} on MEG dataset.}
\label{tab:ablation}
\end{table}

The results demonstrate that each component improves accuracy with little generation time overhead.
Without the shortcut self-consistency loss, the backbone is trained as a standard material SDE, which leads to the largest degradation in accuracy. This confirms that this loss is the primary source of \archname's few-step capability.
Removing the flexible property embedding degrades accuracy mainly at the fewest steps. This aligns with its motivation of accurately recovering unconditioned generation with few steps.
Both backbone mechanisms contribute to keeping the backbone expressive while lightweight, and removing either or both of them degrades accuracy modestly.

Appendix sections~\ref{apx:training-config} and \ref{apx:hp-sensitivity} report \archname's full hyperparameter configuration and sensitivity sweeps.

\section{Conclusion}
This work presents efforts toward the inverse design of amorphous materials and focuses on the inference and training efficiency of generative models for amorphous materials.
We introduce \archname, a generative model capable of performing accurate inverse design of amorphous materials in few sampling steps. It can be trained once and perform generation conditioned on different subsets of target properties. Its denoiser is built on an efficient E(n)-equivariant backbone that captures the complex short- and medium-range order of amorphous materials at a low computational cost. Experiments on three amorphous material datasets provide evidence that \archname{} achieves its design goals.
More discussion can be found in Appendix~\ref{apx:discussion}.

\clearpage

\bibliography{main}

\clearpage
\appendix
\section{Additional Implementation Details}

\subsection{Implementation Details of Material ODE}
\label{apx:material-ode}
Material ODE defines the ground truth $\mu$, $\sigma$, and $\gM_t$ following the optimal transport flow~\cite{DBLP:conf/iclr/LipmanCBNL23}:
\begin{equation}
   \begin{aligned}
      \mu_\mX &= \text{pbc}(\mX_1-\mX_0) \quad \mu_\mE = \mE_1-\mE_0 \\
      \sigma_\mX &= \sigma_\mE = 0 \\
      \mX_t &= \mX_0+t\mu_\mX \quad \mE_t = \mE_0+t\mu_\mE,
   \end{aligned}
\end{equation}
where $\text{pbc}$ denotes the periodic boundary condition, i.e., the vectors between corresponding atoms in $\gM_0$ and $\gM_1$ are adjusted to take the shortest path across periodic boundaries defined by $\mC$.

The network $\mu_\theta$ directly predicts the position and element embedding components of $\mu$, denoted as $\hat \mu_\mX$ and $\hat \mu_\mE$, and is trained with L2 loss. Formally:
\begin{equation}
   \begin{aligned}
      \hat \mu_\mX, \hat \mu_\mE &= \mu_\theta(\gM_t, \vy, t) \\
      \gL_\text{ODE} &= \E_{\gM_0,\gM_1,t} \left[\| \hat \mu_\mX - \mu_\mX \|^2 + \lambda_\mE \| \hat \mu_\mE - \mu_\mE \|^2 \right],
   \end{aligned}
\end{equation}
where $\lambda_\mE$ is the weight that balances the element loss against the position loss, set to $0.5$.
For the random sample $\gM_1$, we sample $\mX_1$ from a uniform distribution within the cell $\mC$, and sample $\mE_1$ from a standard normal distribution.

\subsection{Implementation Details of Material SDE}
\label{apx:material-sde}
Material SDE defines the ground truth following the score-matching SDE~\cite{DBLP:conf/iclr/0011SKKEP21} with a variance exploding noise schedule on positions and a variance preserving noise schedule with cosine progression on element embeddings. For the positions, we have:
\begin{equation}
   \begin{aligned}
      \mu_\mX(\gM_t,t) &= \sigma_\mX^2(t) \scoreX \\
      \sigma_\mX(t) &= t\sigma_\text{max}^\mX \\
      \mX_t &= \mX_0+\sigma_\mX(t)\epsilon_\mX,
   \end{aligned}
   \label{eq:material-sde}
\end{equation}
and for the element embeddings, we have:
\begin{equation}
   \begin{aligned}
      \mu_\mE(\gM_t,t) &= -\frac{\pi}{2}\tan(\pi t/2)\mE_t+\sigma_\mE^2(t) \scoreE \\
      \sigma_\mE(t) &= \sin(\pi t/2)\sigma_\text{max}^\mE\\
      \mE_t &= \cos(\pi t/2)\mE_0+\sigma_\text{max}^\mE\sin(\pi t/2)\epsilon_\mE,
   \end{aligned}
\end{equation}
where $\sigma_\text{max}^\mX$ and $\sigma_\text{max}^\mE$ are the maximum position and element noise scales. $\scoreX$ and $\scoreE$ are the scores of positions and element embeddings, respectively. $\epsilon_\mX$ and $\epsilon_\mE$ are both noise sampled from a standard normal distribution. Under this setting, the distribution of the random sample $\gM_1$ will be consistent with that in material ODE.

The neural network $\mu_\theta$ is set to predict the noise of positions and element embeddings as $\hat \epsilon_\mX$ and $\hat \epsilon_\mE$, and then the predicted scores are $-\hat \epsilon/\sigma(t)$. The network is trained with L2 loss on the noise, formally:
\begin{equation}
   \begin{aligned}
      \hat \epsilon_\mX, \hat \epsilon_\mE &= \mu_\theta(\gM_t, \vy, t) \\
      \gL_\text{SDE} &= \E_{\epsilon,\gM_0,t}\left[ \|\hat\epsilon_\mX - \epsilon_\mX\|^2 + \lambda_\mE \|\hat\epsilon_\mE - \epsilon_\mE\|^2 \right]
   \end{aligned}
\end{equation}

\subsection{Implementation Details of \archname}
\label{apx:mdshortcut}
In our prior experiments, we find that when calculating the shortcut loss $\gL_\text{SC}$ in Eq.~\ref{eq:shortcut-loss}, the stochastic component in material SDE will make the training less stable, leading to suboptimal training results. To solve this, when calculating $\gL_\text{SC}$ we remove the stochastic component in material SDE following the formulation of probabilistic flow ODE introduced in \cite{DBLP:conf/iclr/0011SKKEP21}. Specifically, when calculating Eq.~\ref{eq:shortcut-loss}, $\mu$ and $\sigma$ are defined as:
\begin{equation}
  \begin{aligned}
    \mu_\mX(\gM_t,t) &= \frac{1}{2}\sigma_\mX^2(t) \scoreX\\
    \mu_\mE(\gM_t,t) &= -\frac{\pi}{4}\tan(\pi t/2)\mE_t+\frac{1}{2}\sigma_\mE^2(t) \scoreE\\
    \sigma_\mX &= \sigma_\mE = 0
  \end{aligned}
  \label{eq:probabilistic-ode-flow}
\end{equation}

When a small number $n_s$ of sampling steps is used, the Euler-Maruyama method (Eq.~\ref{eq:euler}) adds a relatively large scale of noise to the sample at each step of the generation process. We find that this will lead to unstable generation results even with \archname. Thus, when $n_s \leq 10$, we follow the probabilistic flow ODE formulation as above for generation.

The shortcut loss is calculated stochastically for only a fraction $\rho_\text{SC}$ of training batches. In preliminary experiments, we find that this is nearly as effective as using all batches, while significantly reducing computational overhead.

\subsection{Implementation Details of the Backbone}
\label{apx:egnn}
The backbone resolves the two length scales that characterize amorphous order.
Short-range order is the bond lengths and bond angles of the first neighbor shells, measured by the ADF, and medium-range order is the network topology measured by the ring size distribution, which reaches up to about $20$\,\AA~\citep{elliott1991medium,salmon2005topological}.

Given an input sample $\gM=(\mC, \mX, \mE)$, the input graph $\gG=(\gV, \gE)$ is composed of the atoms in the sample as nodes in the node set $\gV$, and each edge in the edge set $\gE$ connects a pair of atoms with distance less than a cutoff radius $r_c$. The distances are computed with periodic boundary conditions.
The cutoff radius is chosen to cover the typical first and second neighbor shells of the bonds in the amorphous structures, so that all bonded and strongly interacting atoms share a direct edge connection while the graph stays sufficiently sparse.
The node set also contains a small number $C$ of virtual context nodes that summarize the whole cell. They are phantom nodes that do not correspond to atoms, and each connects to every atom and carries an invariant feature $\mG_c \in \sR^{d_h}$ and $k$ vector channels $\mZ_c \in \sR^{k \times 3}$. Their coordinates are expressed relative to the cell, so that translations, rotations, and reflections of the sample act on them as on the atoms and the exchange stays equivariant.
These context nodes give the backbone a channel for global communication across the cell.

The backbone has $L$ layers.
The $l$-th layer takes as input:
1) node features $\mH^{(l)} \in \sR^{n \times d_h}$ containing information of the corresponding atoms at the $l$-th layer;
2) positional coordinates $\mX^{(l)} \in \sR^{n \times k \times 3}$ of the atoms, where $k$ is the number of vector channels~\cite{levy2023using}; and
3) edge set $\gE$ of the graph $\gG$, with edge attributes $\ve_{ij}$ assigned to each edge.

For the initial layer, the node features $\mH_i^{(0)}$ are assembled by concatenating:
1) diffusion time step $t$;
2) element embeddings: $\mE_i \in \sR^{d_E}$;
3) property embeddings: $\vh_{y_1}, \vh_{y_2}, \ldots, \vh_{y_{n_p}}$ (for each property in $\vy$); and
4) time step size $\Delta t$ for $u_\theta$.
The positions $\mX^{(0)}$ are the original positions $\mX$ replicated across the $k$ channels.
The edge attributes $\ve_{ij}$ are derived from the distance embedding:
\begin{equation}
\ve_{ij} = \tanh\left(\frac{\|\mX_i - \mX_j - \vo_{ij}\|^2}{r_c^2}\right) \cdot 2 - 1,
\end{equation}
where $\vo_{ij}$ is the offset vector accounting for periodic boundary conditions.

Each layer first updates the node feature of every atom, using self-attention~\cite{DBLP:conf/nips/VaswaniSPUJGKP17}:
\begin{equation}
  \begin{aligned}
    \vm_{ij}^{(l)} &= \phi_e^{(l)}(\mH_i^{(l-1)}, \mH_j^{(l-1)}, \ve_{ij}) \\
    \alpha_{ij}^{(l)} &= \sigma(\text{MLP}_{\text{att}}(\vm_{ij}^{(l)})) \\
    \hat{\vm}_{ij}^{(l)} &= \alpha_{ij}^{(l)} \cdot \vm_{ij}^{(l)} \\
    \mH_i^{(l)} &= \mH_i^{(l-1)} + \phi_H^{(l)}\left(\mH_i^{(l-1)}, \sum_{j \in N(i)} \frac{f_\text{cut}(d_{ij}^{(0)}) \cdot \hat{\vm}_{ij}^{(l)}}{n_\text{norm}}\right)
  \end{aligned}
\end{equation}
It then computes a coordinate increment for the vector channels of each atom:
\begin{equation}
  \begin{aligned}
    \mPhi_{ij}^{(l)} &= \text{MLP}_{\text{coord}}([\mH_i^{(l)}, \mH_j^{(l)}, \ve_{ij}]) \in \sR^{k \times k} \\
    \vd_{ij}^{(l)} &= \mX_i^{(l-1)} - \mX_j^{(l-1)} - \vo_{ij} \\
    \mX_i^{(l)}{'} &= \sum_{j \in N(i)} \frac{1}{n_\text{norm}} \cdot \mPhi_{ij}^{(l)} \cdot \vd_{ij}^{(l)},
  \end{aligned}
\end{equation}
where $N(i)$ represents the neighbors of atom $i$, derived from the edge set $\gE$ and
$\sigma$ is the sigmoid activation function for self-attention.
$n_\text{norm}$ is a normalization factor set to the average number of neighbors within the cutoff radius to stabilize the magnitude of the aggregated messages.
$\phi_e^{(l)}$, $\phi_H^{(l)}$ are implemented as multi-layer perceptrons (MLPs) with SiLU activation functions and layer normalization.
$\mPhi_{ij}^{(l)}$ is a learned transformation matrix that maps between the $k$ vector channels.

The vector channels of each atom are then made to interact through their geometry, so that the position update responds to the local arrangement of the neighbors and to the bond angles that pairwise distances alone do not capture.
From the channel increment $\mX_i^{(l)}{'} \in \sR^{k \times 3}$ we form the Gram matrix $\mS_i^{(l)} = \mX_i^{(l)}{'} (\mX_i^{(l)}{'})^\top \in \sR^{k \times k}$, whose entries are the inner products of the channels and are invariant under rotation and reflection, and we collect its upper triangle into a vector $\vs_i^{(l)}$.
We use these invariants to update the node feature and to gate the channels, so the plain residual $\mX_i^{(l)} = \mX_i^{(l-1)} + \mX_i^{(l)}{'}$ becomes:
\begin{equation}
  \begin{aligned}
    \mH_i^{(l)} &\leftarrow \mH_i^{(l)} + \text{MLP}_\text{scal}(\vs_i^{(l)}) \\
    \vg_i^{(l)} &= \sigma\!\left(\text{MLP}_\text{gate}([\mH_i^{(l)}, \vs_i^{(l)}])\right) \in (0,1)^{k} \\
    \mX_i^{(l)} &= \mX_i^{(l-1)} + \vg_i^{(l)} \odot \mX_i^{(l)}{'},
  \end{aligned}
\end{equation}
where $\odot$ broadcasts the per-channel gate over the three spatial axes.
The invariants $\vs_i^{(l)}$ capture the mutual alignment of the channels, hence the angular arrangement of the neighborhood, and the update stays equivariant because $\vs_i^{(l)}$ and $\vg_i^{(l)}$ are invariant while $\mX_i^{(l)}{'}$ is formed from relative positions.
By the first fundamental theorem of invariant theory, the Gram matrix $\mS_i^{(l)}$ is a complete set of invariants of the channels under rotation and reflection~\citep{weyl1946classical,deng2021vector}, so gating the update by a function of $\vs_i^{(l)}$ is a principled invariant modulation.
GVP~\citep{jing2021learning} keeps only the channel norms and PaiNN~\citep{schutt2021equivariant} a single inner product, which are the diagonal of $\mS_i^{(l)}$, whereas the off-diagonal entries used here carry the alignment between distinct channels and thus the angular structure that the diagonal omits.
This scalarization adds angular sensitivity at a cost quadratic in the number of channels $k$, without the spherical harmonics and Clebsch-Gordan tensor products of fully equivariant networks~\citep{thomas2018tensor,batatia2022mace}, whose cost grows steeply with the maximum angular degree~\citep{passaro2023reducing}.
This is what lets the backbone capture the angular short-range order while staying shallow and narrow.

In the same layer, each atom and the context nodes exchange messages with the same edge and channel updates.
With the relative position $\vd_{ci}^{(l)} = \mZ_c^{(l-1)} - \mX_i^{(l)}$ between context node $c$ and atom $i$, and an edge attribute $\ve_{ci}$ derived from its length as for the atom edges, each atom is updated from the context nodes by:
\begin{equation}
  \begin{aligned}
    \vm_{ci}^{(l)} &= \phi_e^{(l)}(\mH_i^{(l)}, \mG_c^{(l-1)}, \ve_{ci}) \\
    \bm{\Psi}_{ci}^{(l)} &= \text{MLP}_{\text{coord}}([\mH_i^{(l)}, \mG_c^{(l-1)}, \ve_{ci}]) \in \sR^{k \times k} \\
    \mH_i^{(l)} &\leftarrow \mH_i^{(l)} + \phi_H^{(l)}\left(\mH_i^{(l)}, \frac{1}{C}\sum_{c} \vm_{ci}^{(l)}\right) \\
    \mX_i^{(l)} &\leftarrow \mX_i^{(l)} - \frac{1}{C}\sum_{c} \bm{\Psi}_{ci}^{(l)} \vd_{ci}^{(l)},
  \end{aligned}
\end{equation}
and each context node is updated from the atoms by:
\begin{equation}
  \begin{aligned}
    \mG_c^{(l)} &= \mG_c^{(l-1)} + \phi_H^{(l)}\left(\mG_c^{(l-1)}, \frac{1}{n}\sum_{i} \vm_{ci}^{(l)}\right) \\
    \mZ_c^{(l)} &= \mZ_c^{(l-1)} + \frac{1}{n}\sum_{i} \bm{\Psi}_{ci}^{(l)} \vd_{ci}^{(l)},
  \end{aligned}
\end{equation}
where $C$ is the number of context nodes and $n$ the number of atoms, so each atom reads a summary of the whole cell while each context node reads the atoms.
This carries such a summary to every atom within a single layer, which a local graph would reach only after many layers, on the order of the cell width divided by the cutoff radius~\citep{di2023over,sanchez2020learning}, and it stays equivariant because the updates act through the relative positions $\vd_{ci}^{(l)}$ while the messages $\vm_{ci}^{(l)}$ and the mixing matrices $\bm{\Psi}_{ci}^{(l)}$ are invariant.
The context nodes thereby provide global communication at a cost linear in the number of atoms, rather than the quadratic cost of attention between every pair of atoms~\citep{cai2023connection}.
After the last layer the context nodes are discarded, and only the atoms are read out.

The MLPs above are structured as follows:
\begin{equation}
  \begin{aligned}
    \phi_e^{(l)}(\mH_i, \mH_j, \ve_{ij}) &= \text{MLP}_{\text{edge}}([\mH_i, \mH_j, \ve_{ij}]) \\
    \phi_H^{(l)}(\mH_i, \vm_{\text{agg}}) &= \text{MLP}_{\text{node}}([\mH_i, \vm_{\text{agg}}]),
  \end{aligned}
\end{equation}
where each MLP uses SiLU activation functions and layer normalization.
A smooth cutoff function $f_\text{cut}$ is used to prevent discontinuities when atoms leave or enter the cutoff radius:
\begin{equation}
  f_\text{cut}(r) = 2 \tanh\left(1 - \frac{\min(r, r_c)}{r_c}\right)^2
\end{equation}

At the last layer, the backbone outputs $\mH^{(L)}$ and $\mX^{(L)}$ as the final node features and positional coordinates, respectively. We take $\mH^{(L)}$ directly as the predicted element component, and the deviation between the original positions and the first channel of output positions $\mX - \mX^{(L,0)}$ as the predicted position component.

\subsection{Hyperparameter Configuration}
\label{apx:training-config}
Table~\ref{tab:hyperparams} lists the full set of chosen hyperparameters of \archname.
For every baseline, we use the optimal configuration reported in its corresponding publication.

\begin{table}[h]
\centering
\begin{tabular}{ll}
\toprule
Hyperparameter & Value \\
\midrule
\multicolumn{2}{l}{\textbf{E(n)-equivariant backbone}} \\
\quad Number of layers $L$ & 3 \\
\quad Hidden dimension $d_h$ & 64 \\
\quad Vector channels $k$ & 8 \\
\quad Neighbor cutoff $r_c$ & 6.5\,\AA \\
\quad Normalization factor $n_\text{norm}$ & 40 \\
\quad Self-attention hidden dim & 64 \\
\quad Number of context nodes $C$ & 4 \\
\midrule
\multicolumn{2}{l}{\textbf{Material SDE}} \\
\quad Position noise scale $\sigma_{\max}^{\mX}$ & 1.7\,\AA \\
\quad Element noise scale $\sigma_{\max}^{\mE}$ & 1.5 \\
\quad Element loss weight $\lambda_{\mE}$ & 0.5 \\
\midrule
\multicolumn{2}{l}{\textbf{Training}} \\
\quad Shortcut-loss batch fraction $\rho_\text{SC}$ & 25\% \\
\quad Property dropout probability $p_\text{drop}$ & 0.2 \\
\quad Optimizer & Adam \\
\quad Learning rate & $10^{-3}$ \\
\quad Number of epochs & 500 \\
\quad Batch size & 8 \\
\bottomrule
\end{tabular}

\caption{Hyperparameters of \archname.}
\label{tab:hyperparams}
\end{table}

\subsection{Structural Metrics Calculation}
\label{apx:structural-metrics}

\paragraph{Radial distribution function (RDF).}
The RDF $g(r)$ quantifies the probability of finding an atom at distance $r$ from a reference atom, normalized by the corresponding probability in an ideal gas:
\begin{equation}
g(r) = \frac{V}{N^2} \frac{1}{4\pi r^2 \Delta r} \left\langle \sum_{i=1}^{N} \sum_{j \neq i}^{N} \delta(r - r_{ij}) \right\rangle,
\end{equation}
where $V$ is the system volume, $N$ is the number of atoms, $r_{ij}$ is the distance between atoms $i$ and $j$, and $\langle \cdot \rangle$ denotes ensemble averaging. We calculate RDF using the ASE~\citep{larsen2017atomic} library's \texttt{Analysis} module with a cutoff distance of 5.0 \AA{} and 100 bins. Periodic boundary conditions are applied to ensure proper treatment of atoms near cell boundaries.

\paragraph{Angular distribution function (ADF).}
The ADF $P(\theta)$ measures the distribution of bond angles formed by atomic triplets. For each central atom $i$, we identify all neighbors $j$ and $k$ within a cutoff radius of 3.0 \AA, then calculate the angle $\theta_{jik}$ between vectors $\vr_{ij}$ and $\vr_{ik}$:
\begin{equation}
\cos(\theta_{jik}) = \frac{\vr_{ij} \cdot \vr_{ik}}{|\vr_{ij}||\vr_{ik}|}
\end{equation}
The distribution is binned into 180 bins covering 0$^\circ$ to 180$^\circ$ (1$^\circ$ resolution). Neighbor lists are constructed using ASE's \texttt{NeighborList} with periodic boundary conditions.

\paragraph{RMSD Calculation.}
The structural accuracy is quantified using RMSD between generated and reference distributions:
\begin{equation}
\text{RMSD} = \sqrt{\frac{1}{N_{\text{bins}}} \sum_{i=1}^{N_{\text{bins}}} (f_i^{\text{gen}} - f_i^{\text{ref}})^2},
\end{equation}
where $f_i$ represents the distribution value at bin $i$. For multiple samples, we first average the distributions across all structures before computing RMSD.

\subsection{Property Calculation}
\label{apx:property-calculation}
We evaluate inverse design performance using material properties relevant to the dataset. For a-\ch{SiO2} samples, we focus on shear modulus and ring size distribution. For MEG samples, we focus on Young's modulus, shear modulus, and lithium molar concentration.

\paragraph{a-\ch{SiO2} dataset properties.}
For a-\ch{SiO2} samples, properties are computed directly from atomic structures. Shear modulus is calculated using finite differences of the stress tensor. Structures are relaxed with the Tersoff potential (force tolerance 0.05 eV/\AA), then subjected to small strains ($\delta = 0.02$) to compute elastic constants $C_{44}$, $C_{55}$, and $C_{66}$ from stress responses. Shear modulus is their average: $G = \frac{1}{3}(C_{44} + C_{55} + C_{66})$. Ring size distribution is computed by identifying closed rings in the Si-O network using a depth-first search algorithm. Atoms are considered bonded if their distance is below 1.3 times the sum of their covalent radii (Si: 1.11 \AA, O: 0.66 \AA). The algorithm searches for the shortest path between bonded atom pairs while excluding the direct bond, ensuring proper ring closure with periodic boundary conditions. Ring sizes are reported as the number of Si atoms per ring, averaged across all identified rings.

\paragraph{MEG dataset properties.}
For MEG samples, elastic properties are computed using the BMP-shrm potential~\cite{bertani2021improved}. Structures are first relaxed, then elastic constants $C_{ij}$ are calculated from finite differences of the stress tensor using Voigt notation. Young's modulus is computed as $E = \frac{(C_{11} - C_{12})(C_{11} + 2C_{12})}{C_{11} + C_{12}}$ and shear modulus as $G = C_{44}$. Lithium molar concentration $C_\text{Li}$ is computed directly from the elemental composition as the ratio of lithium atoms to total atoms.

\paragraph{Implementation details.}
Ghost atoms are excluded from all calculations. Structural relaxations use quasi-Newton optimization with variable cell shapes, maximum 1500 steps. Ring finding incorporates periodic boundary conditions for rings crossing cell boundaries.

\subsection{Inverse Design Metrics}
\label{apx:inverse-design-metrics}

For inverse design evaluation, we compare target properties $y_{\text{target}}$ with computed or predicted properties $y_{\text{generated}}$ of generated samples using three standard regression metrics:

\paragraph{Mean Absolute Error (MAE).}
\begin{equation}
\text{MAE} = \frac{1}{n} \sum_{i=1}^{n} |y_{\text{generated},i} - y_{\text{target},i}|
\end{equation}

\paragraph{Root Mean Square Error (RMSE).}
\begin{equation}
\text{RMSE} = \sqrt{\frac{1}{n} \sum_{i=1}^{n} (y_{\text{generated},i} - y_{\text{target},i})^2}
\end{equation}

\paragraph{Mean Absolute Percentage Error (MAPE).}
\begin{equation}
\text{MAPE} = \frac{100\%}{n} \sum_{i=1}^{n} \left|\frac{y_{\text{generated},i} - y_{\text{target},i}}{y_{\text{target},i}}\right|
\end{equation}

These metrics are computed separately for each property (shear modulus, RSD, Young's modulus, Li ratio) across all generated samples. MAE provides interpretable error magnitude in original units, RMSE penalizes large deviations more heavily, and MAPE enables comparison across properties with different scales and units.

\subsection{Details on Dataset Preparation}
\label{apx:datasets}

\subsubsection{Amorphous Silicon (a-Si) Dataset}
The a-Si dataset contains $10\,000$ samples created using LAMMPS~\cite{lammps} software with the Stillinger--Weber potential~\cite{stillinger1985computer}. All molecular dynamics (MD) simulations are performed in the NPT ensemble at zero pressure.
The dataset is generated by heating the crystalline silicon from 2500\,K to 3000\,K over 200\,ps, equilibrating the melt for 300\,ps, and then cooling it down again to 2500\,K at a rate of $10^{12}$\,K/s.
The final samples are taken after equilibrating for another 300\,ps at 2500\,K.

\subsubsection{Amorphous Silica (a-\ch{SiO2}) Dataset}
The a-\ch{SiO2} dataset contains $6,000$ samples that share the composition of pure silica, \ch{SiO2}.
To maximize the variation of properties between the samples generated with the same simulation workflow, relatively small unit cells are chosen with the number of atoms uniformly selected in the range of 80 to 250.
Atoms are initially placed in a unit cell with a volume $V = 4 \sum_i \frac{4}{3} \pi r_i^3$, with $r_i$ being the covalent radius of the $i$-th atom, avoiding unphysical overlap between neighboring atoms.
A local structure relaxation is performed on the initial configuration followed by an MD simulation in the NPT ensemble at 3500\,K for 2000\,ps.
To limit the effects of relaxation, which we observe in our other datasets, we use an instantaneous quenching procedure by performing a local structure optimization and a subsequent equilibration at 300\,K for 10\,ps.
All simulations are performed using LAMMPS~\cite{lammps} software and the Tersoff potential parameterized by \cite{munetoh2007interatomic}.

\subsubsection{Multi-Element Glass (MEG) Dataset}
The MEG dataset contains 9,027 multi-component glass samples with 11 different elements (Si, P, Al, Li, Be, K, Ca, Ti, Ba, Zn, O).
Compositions are generated from varying ratios of glass formers (\ch{SiO2}, \ch{P2O5}) and modifiers (\ch{Al2O3}, \ch{Li2O}, \ch{BeO}, \ch{K2O}, \ch{CaO}, \ch{TiO2}, \ch{BaO}, \ch{ZnO}), with up to four modifiers at 40\% total concentration relative to the glass formers.
Initial structures contain approximately 800 atoms placed randomly in a simulation cell with volume $V = 3 \sum_i \frac{4}{3} \pi r_i^3$, where $r_i$ is the covalent radius of atom $i$.
Samples are prepared using a melt-quench procedure: melting at $\frac{3}{4} T_\mathrm{evap}$ for 400\,ps, quenching to 300\,K at 5\,K/ps, and equilibrating for 300\,ps.
All simulations use LAMMPS~\cite{lammps} with the Bertani--Menziani--Pedone (BMP)-shrm potential~\cite{bertani2021improved}.

\subsection{Experimental Environment}
\label{apx:environment}

\paragraph{Hardware.}
All experiments are run on an internal GPU cluster managed by the SLURM scheduler.
Each compute node is equipped with two Intel Xeon Platinum 8168 24-core processors (48 cores, 96 threads total, base 2.7\,GHz, boost 3.7\,GHz), 1.5\,TB of system memory, and NVIDIA Tesla V100-SXM3 GPUs with 32\,GB of memory each, running driver version 580.159.03.
Each experiment uses a single V100 GPU.

\paragraph{Software.}
The implementation uses Python~3.10 and PyTorch~2.6.0~\citep{paszke2019pytorch} built against CUDA~12.6.
Key dependencies include NumPy~\citep{harris2020array} and ASE~\citep{larsen2017atomic}, with atomic neighbor lists built through torch-nl.
All experiments run inside a Singularity~\citep{kurtzer2017singularity} container built from the \texttt{ubuntu:22.04} base image to ensure a reproducible software environment across compute nodes.

\paragraph{Evaluation protocol.}
Every metric we report is the mean over 3 repetitions of the corresponding experiment.
We do not fix the random seed and leave it to PyTorch's default random behavior, so the five repetitions differ in their random initialization and data ordering.

\section{Additional Experimental Results}

\begin{figure*}[t]
    \centering
    \begin{subfigure}[b]{1.0\linewidth}
        \centering
        \includegraphics[width=0.242\linewidth]{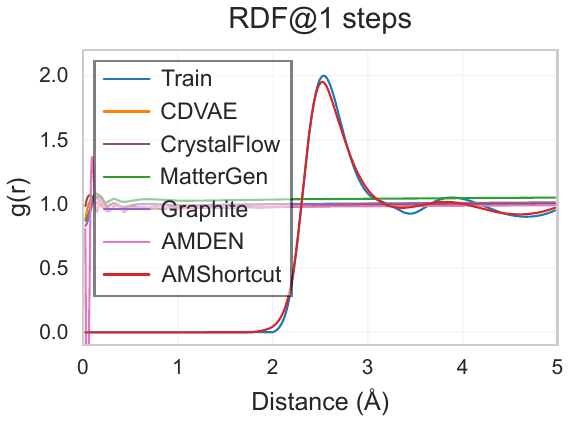}
        \includegraphics[width=0.242\linewidth]{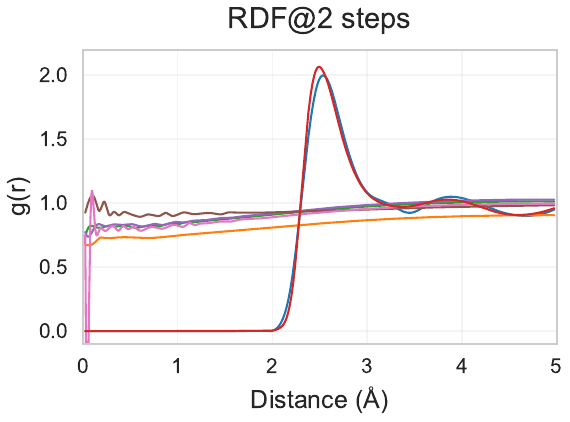}
        \includegraphics[width=0.242\linewidth]{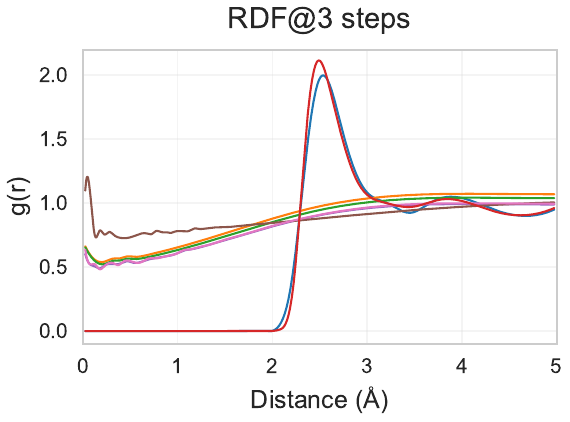}
        \includegraphics[width=0.242\linewidth]{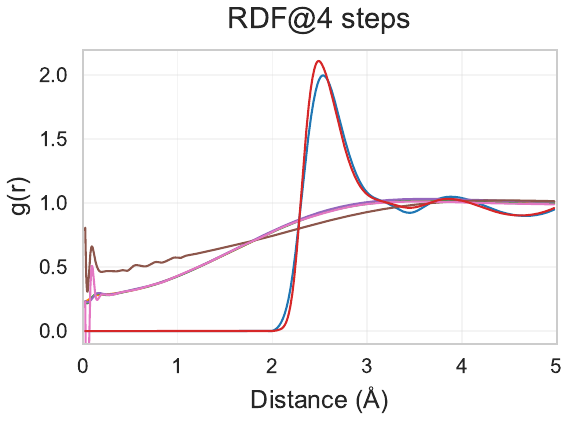}\\
        \includegraphics[width=0.242\linewidth]{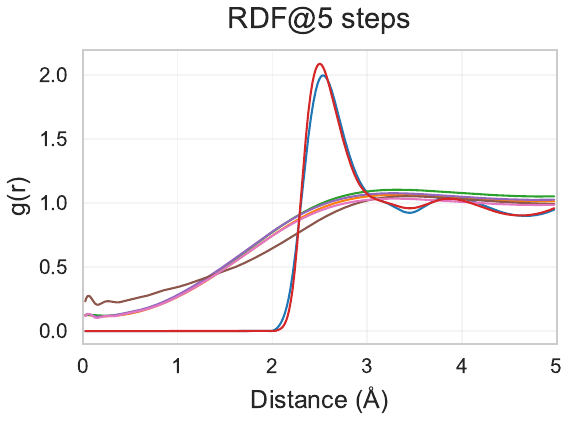}
        \includegraphics[width=0.242\linewidth]{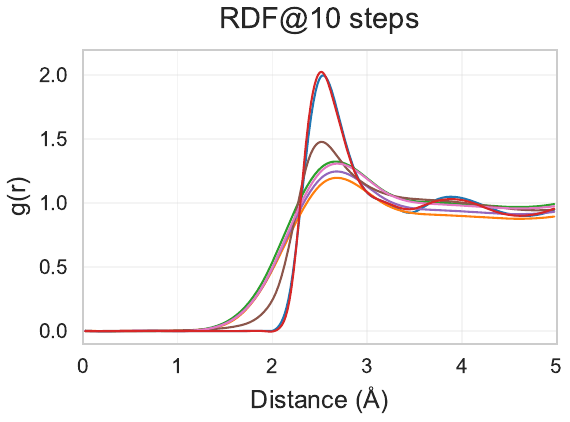}
        \includegraphics[width=0.242\linewidth]{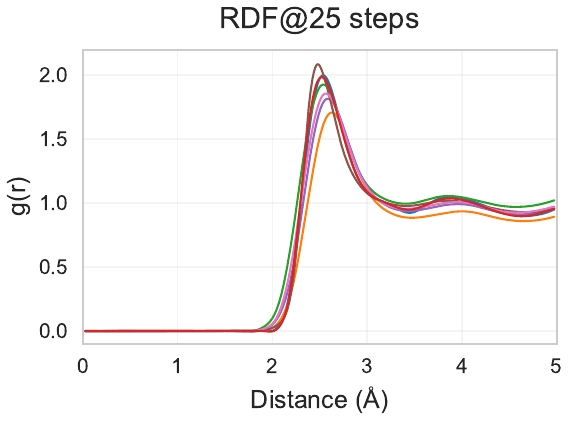}
        \includegraphics[width=0.242\linewidth]{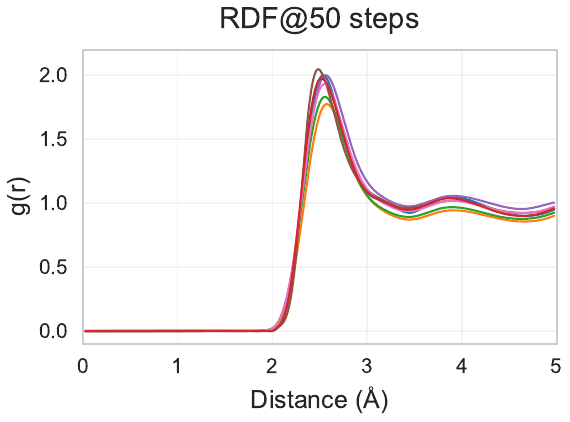}
        \caption{Radial distribution function (RDF)}
    \end{subfigure}

    \vspace{10pt}

    \begin{subfigure}[b]{1.0\linewidth}
        \centering
        \includegraphics[width=0.242\linewidth]{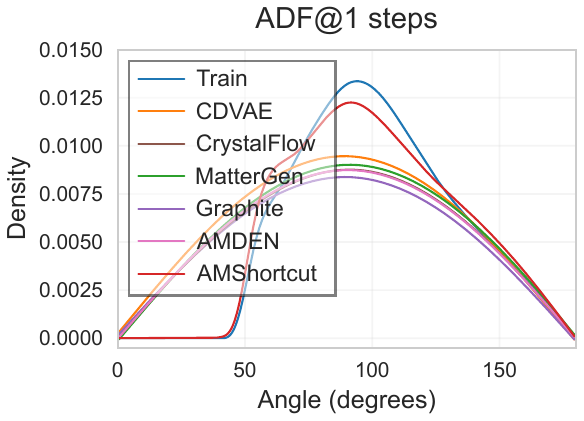}
        \includegraphics[width=0.242\linewidth]{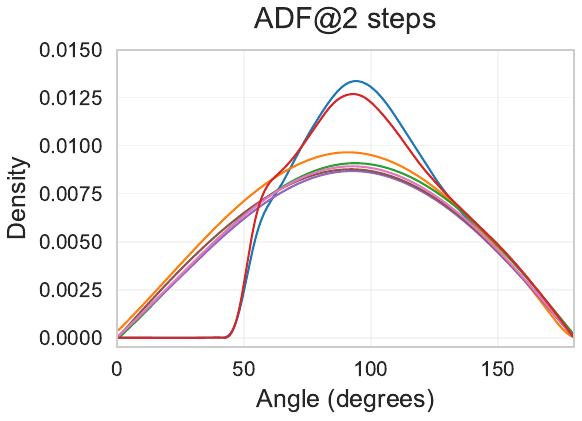}
        \includegraphics[width=0.242\linewidth]{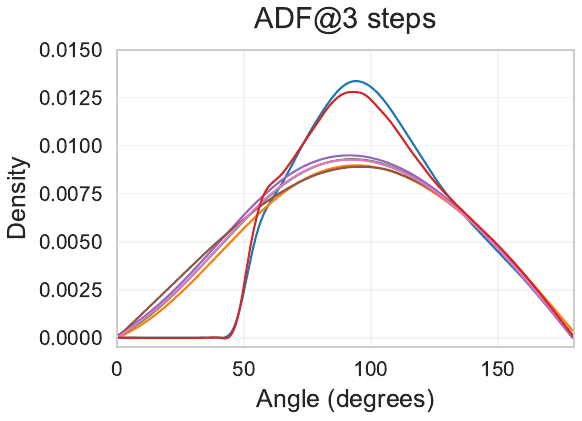}
        \includegraphics[width=0.242\linewidth]{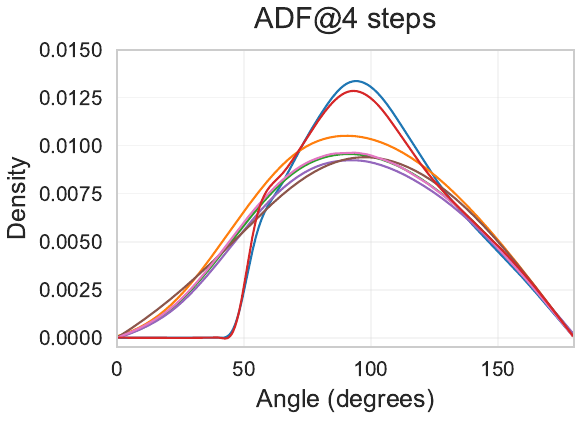}\\
        \includegraphics[width=0.242\linewidth]{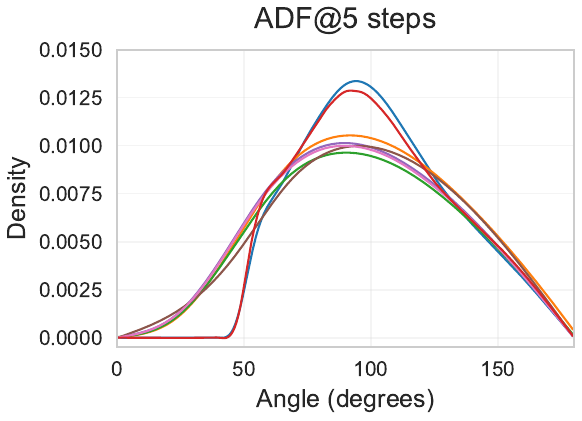}
        \includegraphics[width=0.242\linewidth]{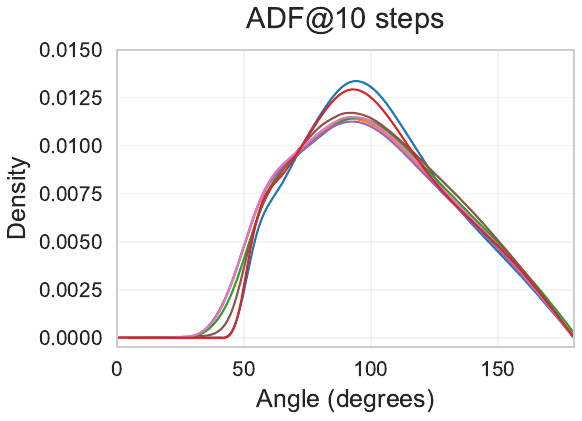}
        \includegraphics[width=0.242\linewidth]{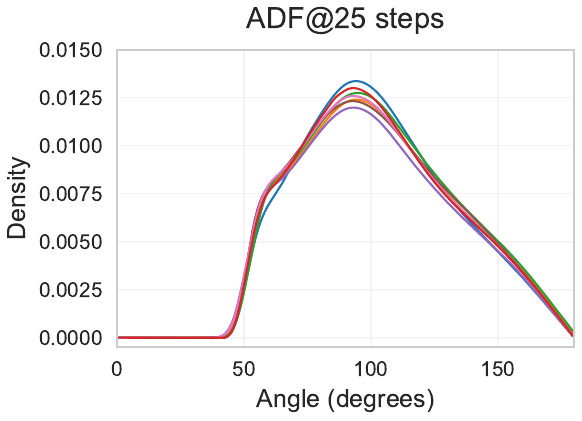}
        \includegraphics[width=0.242\linewidth]{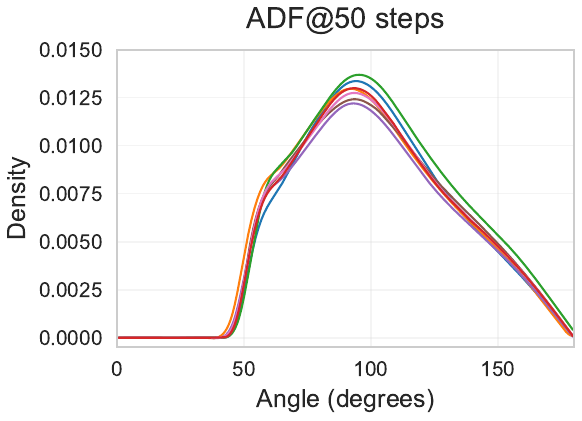}
        \caption{Angular distribution function (ADF)}
    \end{subfigure}
    \caption{Comparison of structural functions on the a-\ch{Si} dataset.}
    \label{fig:structural-accuracy}
\end{figure*}

\subsection{RDF and ADF Plots of Models on a-\ch{Si} Dataset}
\label{apx:rdf-adf}
Figure~\ref{fig:structural-accuracy} shows the RDF and ADF of samples generated by different models across various sampling step counts, demonstrating that \archname{} achieves structural accuracy comparable to training samples even with few steps.

\begin{figure*}[t]
    \centering
    \begin{subfigure}[b]{0.48\linewidth}
        \centering

\resizebox{\linewidth}{!}{%
\begin{tikzpicture}
\begin{axis}[
  scatteraxis,
  xmode=log, log basis x=10,
  xlabel={Generation Time per 10,000 samples (s)},
  ylabel={RDF RMSD},
  xmin=60, xmax=18000, ymin=-0.02, ymax=0.74,
  xtick={100,1000,10000},
  ytick={0,0.1,0.2,0.3,0.4,0.5,0.6,0.7},
  legend pos=north east,
]
\addplot[scpts, sCDVAE] coordinates {
    (84.0, 0.69442)  [1]
    (138.0, 0.57868)  [2]
    (198.0, 0.53362)  [3]
    (264.0, 0.4345)  [4]
    (264.0, 0.39258)  [5]
    (528.0, 0.21683)  [10]
    (1362.0, 0.08542)  [25]
    (2784.0, 0.08517)  [50]
    (5760.0, 0.11667)  [100]
    (12960.0, 0.07572)  [250]
};
\addlegendentry{CDVAE}
\addplot[scpts, sODE] coordinates {
    (105.9, 0.69118)  [1]
    (134.2, 0.66442)  [2]
    (169.3, 0.60209)  [3]
    (222.4, 0.49544)  [4]
    (277.6, 0.39007)  [5]
    (545.8, 0.14917)  [10]
    (1347.5, 0.04876)  [25]
    (2695.3, 0.04203)  [50]
    (5406.5, 0.03305)  [100]
    (12163.6, 0.03353)  [250]
};
\addlegendentry{CrystalFlow}
\addplot[scpts, sMatterGen] coordinates {
    (96.0, 0.67755)  [1]
    (150.0, 0.60909)  [2]
    (210.0, 0.5054)  [3]
    (306.0, 0.43985)  [4]
    (324.0, 0.38154)  [5]
    (600.0, 0.22949)  [10]
    (1608.0, 0.09168)  [25]
    (3162.0, 0.0763)  [50]
    (6120.0, 0.02898)  [100]
    (14760.0, 0.0439)  [250]
};
\addlegendentry{MatterGen}
\addplot[scpts, sGraphite] coordinates {
    (72.0, 0.69555)  [1]
    (114.0, 0.61882)  [2]
    (180.0, 0.51208)  [3]
    (228.0, 0.43101)  [4]
    (246.0, 0.38436)  [5]
    (492.0, 0.22334)  [10]
    (1266.0, 0.05719)  [25]
    (2418.0, 0.04006)  [50]
    (4680.0, 0.03667)  [100]
    (11880.0, 0.0334)  [250]
};
\addlegendentry{Graphite}
\addplot[scpts, sSDE] coordinates {
    (72.5, 0.68477)  [1]
    (117.4, 0.60972)  [2]
    (171.9, 0.50904)  [3]
    (227.9, 0.43276)  [4]
    (238.3, 0.37975)  [5]
    (475.0, 0.22054)  [10]
    (1210.5, 0.04871)  [25]
    (2387.6, 0.02631)  [50]
    (4806.2, 0.02143)  [100]
    (11552.5, 0.01905)  [250]
};
\addlegendentry{AMDEN}
\addplot[scpts, sShortcut] coordinates {
    (69.0, 0.02513)  [1]
    (110.0, 0.03922)  [2]
    (165.9, 0.04835)  [3]
    (183.1, 0.04653)  [4]
    (206.4, 0.04041)  [5]
    (304.9, 0.02244)  [10]
    (637.9, 0.01618)  [25]
    (1208.8, 0.01487)  [50]
    (2430.9, 0.0158)  [100]
    (5238.2, 0.01548)  [250]
};
\addlegendentry{\archname{}}
\end{axis}
\end{tikzpicture}%
}
        \vspace{-10pt}
        \caption{Radial distribution function (RDF)}
    \end{subfigure}
    \hfill
    \begin{subfigure}[b]{0.48\linewidth}
        \centering

\resizebox{\linewidth}{!}{%
\begin{tikzpicture}
\begin{axis}[
  scatteraxis,
  xmode=log, log basis x=10,
  xlabel={Generation Time per 10,000 samples (s)},
  ylabel={ADF RMSD},
  xmin=60, xmax=18000, ymin=0.0002, ymax=0.00315,
  xtick={100,1000,10000},
  ytick={0.0005,0.0010,0.0015,0.0020,0.0025,0.0030},
  scaled y ticks=false, yticklabel style={/pgf/number format/fixed, /pgf/number format/precision=4},
  legend pos=north east,
]
\addplot[scpts, sCDVAE] coordinates {
    (84.0, 0.00265)  [1]
    (138.0, 0.00265)  [2]
    (198.0, 0.00251)  [3]
    (264.0, 0.00203)  [4]
    (264.0, 0.00173)  [5]
    (528.0, 0.00137)  [10]
    (1362.0, 0.00048)  [25]
    (2784.0, 0.00043)  [50]
    (5760.0, 0.0008)  [100]
    (12960.0, 0.00077)  [250]
};
\addlegendentry{CDVAE}
\addplot[scpts, sODE] coordinates {
    (105.9, 0.0028)  [1]
    (134.2, 0.00276)  [2]
    (169.3, 0.00259)  [3]
    (222.4, 0.0022)  [4]
    (277.6, 0.00173)  [5]
    (545.8, 0.00073)  [10]
    (1347.5, 0.00051)  [25]
    (2695.3, 0.00048)  [50]
    (5406.5, 0.00046)  [100]
    (12163.6, 0.00045)  [250]
};
\addlegendentry{CrystalFlow}
\addplot[scpts, sMatterGen] coordinates {
    (96.0, 0.00286)  [1]
    (150.0, 0.00272)  [2]
    (210.0, 0.00231)  [3]
    (306.0, 0.00221)  [4]
    (324.0, 0.00175)  [5]
    (600.0, 0.00094)  [10]
    (1608.0, 0.00069)  [25]
    (3162.0, 0.00041)  [50]
    (6120.0, 0.00049)  [100]
    (14760.0, 0.00073)  [250]
};
\addlegendentry{MatterGen}
\addplot[scpts, sGraphite] coordinates {
    (72.0, 0.00279)  [1]
    (114.0, 0.00264)  [2]
    (180.0, 0.00242)  [3]
    (228.0, 0.00203)  [4]
    (246.0, 0.00198)  [5]
    (492.0, 0.00104)  [10]
    (1266.0, 0.00055)  [25]
    (2418.0, 0.00034)  [50]
    (4680.0, 0.0004)  [100]
    (11880.0, 0.00053)  [250]
};
\addlegendentry{Graphite}
\addplot[scpts, sSDE] coordinates {
    (72.5, 0.00282)  [1]
    (117.4, 0.00263)  [2]
    (171.9, 0.00235)  [3]
    (227.9, 0.0021)  [4]
    (238.3, 0.00188)  [5]
    (475.0, 0.001)  [10]
    (1210.5, 0.00045)  [25]
    (2387.6, 0.00039)  [50]
    (4806.2, 0.00037)  [100]
    (11552.5, 0.00035)  [250]
};
\addlegendentry{AMDEN}
\addplot[scpts, sShortcut] coordinates {
    (69.0, 0.00067)  [1]
    (110.0, 0.00044)  [2]
    (165.9, 0.00038)  [3]
    (183.1, 0.00035)  [4]
    (206.4, 0.00034)  [5]
    (304.9, 0.00032)  [10]
    (637.9, 0.00029)  [25]
    (1208.8, 0.0003)  [50]
    (2430.9, 0.00028)  [100]
    (5238.2, 0.00029)  [250]
};
\addlegendentry{\archname{}}
\end{axis}
\end{tikzpicture}%
}
        \vspace{-10pt}
        \caption{Angular distribution function (ADF)}
    \end{subfigure}
    \caption{Structural metrics versus generation time per 10,000 a-\ch{Si} samples. Labels indicate the number of sampling steps.}
    \label{fig:structural-accuracy-vs-time}
\end{figure*}

\begin{figure*}[t]
    \centering
    \begin{subfigure}[b]{0.48\linewidth}
        \centering

\resizebox{\linewidth}{!}{%
\begin{tikzpicture}
\begin{axis}[
  scatteraxis,
  xmode=log, log basis x=10,
  xlabel={Generation Time per 2,000 samples (s)},
  ylabel={MAPE (\%)},
  xmin=140, xmax=60000, ymin=8, ymax=93,
  xtick={1000,10000},
  ytick={10,20,30,40,50,60,70,80,90},
  legend pos=north east,
]
\addplot[scpts, sCDVAE] coordinates {
    (342.7, 72.3)  [1]
    (987.9, 68.2)  [5]
    (1803.4, 74.8)  [10]
    (4231.5, 18.9)  [25]
    (8306.4, 17.1)  [50]
    (16381.8, 15.8)  [100]
    (40599.1, 15.1)  [250]
};
\addlegendentry{CDVAE}
\addplot[scpts, sODE] coordinates {
    (233.4, 68.7)  [1]
    (713.8, 52.0)  [5]
    (1421.9, 33.9)  [10]
    (3595.4, 20.1)  [25]
    (7026.5, 17.3)  [50]
    (14101.7, 16.0)  [100]
    (35203.7, 14.9)  [250]
};
\addlegendentry{CrystalFlow}
\addplot[scpts, sMatterGen] coordinates {
    (387.4, 62.8)  [1]
    (1116.7, 62.1)  [5]
    (2038.7, 61.3)  [10]
    (4783.5, 16.4)  [25]
    (9389.8, 15.6)  [50]
    (18518.6, 14.9)  [100]
    (45894.6, 14.5)  [250]
};
\addlegendentry{MatterGen}
\addplot[scpts, sGraphite] coordinates {
    (298.0, 52.9)  [1]
    (859.0, 51.2)  [5]
    (1568.2, 56.8)  [10]
    (3679.6, 15.2)  [25]
    (7222.9, 14.9)  [50]
    (14245.1, 14.3)  [100]
    (35303.6, 14.1)  [250]
};
\addlegendentry{Graphite}
\addplot[scpts, sSDE] coordinates {
    (298.0, 54.8)  [1]
    (859.0, 80.5)  [5]
    (1568.2, 65.8)  [10]
    (3679.6, 14.7)  [25]
    (7222.9, 14.7)  [50]
    (14245.1, 15.2)  [100]
    (35303.6, 15.6)  [250]
};
\addlegendentry{AMDEN}
\addplot[scpts, sShortcut] coordinates {
    (224.6, 20.5)  [1]
    (437.2, 14.0)  [5]
    (711.1, 13.0)  [10]
    (1509.4, 12.7)  [25]
    (2856.8, 12.7)  [50]
    (5990.9, 12.9)  [100]
    (13889.9, 13.0)  [250]
};
\addlegendentry{\archname{}}
\end{axis}
\end{tikzpicture}%
}
        \vspace{-10pt}
        \caption{Shear Modulus [GPa]}
    \end{subfigure}
    \hfill
    \begin{subfigure}[b]{0.48\linewidth}
        \centering

\resizebox{\linewidth}{!}{%
\begin{tikzpicture}
\begin{axis}[
  scatteraxis,
  xmode=log, log basis x=10,
  xlabel={Generation Time per 2,000 samples (s)},
  ylabel={MAPE (\%)},
  xmin=120, xmax=60000, ymin=1, ymax=53,
  xtick={1000,10000},
  ytick={10,20,30,40,50},
  legend pos=north east,
]
\addplot[scpts, sCDVAE] coordinates {
    (342.7, 49.8)  [1]
    (987.9, 48.9)  [5]
    (1803.4, 48.3)  [10]
    (4231.5, 5.7)  [25]
    (8306.4, 5.2)  [50]
    (16381.8, 4.5)  [100]
    (40599.1, 4.3)  [250]
};
\addlegendentry{CDVAE}
\addplot[scpts, sODE] coordinates {
    (233.4, 43.2)  [1]
    (713.8, 37.1)  [5]
    (1421.9, 7.3)  [10]
    (3595.4, 4.0)  [25]
    (7026.5, 4.2)  [50]
    (14101.7, 4.3)  [100]
    (35203.7, 4.3)  [250]
};
\addlegendentry{CrystalFlow}
\addplot[scpts, sMatterGen] coordinates {
    (387.4, 44.3)  [1]
    (1116.7, 43.1)  [5]
    (2038.7, 42.6)  [10]
    (4783.5, 4.6)  [25]
    (9389.8, 4.4)  [50]
    (18518.6, 3.9)  [100]
    (45894.6, 3.8)  [250]
};
\addlegendentry{MatterGen}
\addplot[scpts, sGraphite] coordinates {
    (298.0, 41.8)  [1]
    (859.0, 40.6)  [5]
    (1568.2, 40.2)  [10]
    (3679.6, 4.0)  [25]
    (7222.9, 3.9)  [50]
    (14245.1, 3.6)  [100]
    (35303.6, 3.6)  [250]
};
\addlegendentry{Graphite}
\addplot[scpts, sSDE] coordinates {
    (298.0, 43.2)  [1]
    (859.0, 37.9)  [5]
    (1568.2, 43.9)  [10]
    (3679.6, 3.9)  [25]
    (7222.9, 3.8)  [50]
    (14245.1, 3.5)  [100]
    (35303.6, 3.6)  [250]
};
\addlegendentry{AMDEN}
\addplot[scpts, sShortcut] coordinates {
    (224.6, 4.6)  [1]
    (437.2, 3.3)  [5]
    (711.1, 3.0)  [10]
    (1509.4, 2.8)  [25]
    (2856.8, 2.9)  [50]
    (5990.9, 3.0)  [100]
    (13889.9, 3.0)  [250]
};
\addlegendentry{\archname{}}
\end{axis}
\end{tikzpicture}%
}
        \vspace{-10pt}
        \caption{Ring size distribution (RSD)}
    \end{subfigure}
    \caption{Inverse design metrics versus generation time per 2,000 a-\ch{SiO2} samples. Labels indicate the number of sampling steps.}
    \label{fig:sio2-inverse-accuracy-vs-time}
\end{figure*}

\begin{figure*}[t]
    \centering
    \begin{subfigure}[b]{0.32\linewidth}
        \centering

\resizebox{\linewidth}{!}{%
\begin{tikzpicture}
\begin{axis}[
  scatteraxis,
  xmode=log, log basis x=10,
  xlabel={Generation Time per 2,000 samples (s)},
  ylabel={MAPE (\%)},
  xmin=240, xmax=130000, ymin=14, ymax=72,
  xtick={1000,10000,100000},
  ytick={20,30,40,50,60,70},
  legend pos=north east,
]
\addplot[scpts, sCDVAE] coordinates {
    (689.5, 68.2)  [1]
    (2024.6, 62.5)  [5]
    (3635.8, 49.5)  [10]
    (8494.6, 29.7)  [25]
    (16620.4, 27.7)  [50]
    (32992.2, 26.3)  [100]
    (81742.6, 25.1)  [250]
};
\addlegendentry{CDVAE}
\addplot[scpts, sODE] coordinates {
    (321.4, 66.5)  [1]
    (1438.4, 56.7)  [5]
    (2852.3, 40.8)  [10]
    (7091.5, 24.6)  [25]
    (14149.8, 23.4)  [50]
    (28352.4, 22.8)  [100]
    (70565.0, 22.6)  [250]
};
\addlegendentry{CrystalFlow}
\addplot[scpts, sMatterGen] coordinates {
    (779.5, 63.8)  [1]
    (2288.7, 58.8)  [5]
    (4110.0, 47.5)  [10]
    (9602.6, 26.4)  [25]
    (18788.3, 24.2)  [50]
    (37295.6, 22.9)  [100]
    (92404.7, 21.9)  [250]
};
\addlegendentry{MatterGen}
\addplot[scpts, sGraphite] coordinates {
    (599.6, 58.5)  [1]
    (1760.5, 55.8)  [5]
    (3161.5, 44.8)  [10]
    (7386.6, 23.1)  [25]
    (14452.5, 21.5)  [50]
    (28688.9, 20.5)  [100]
    (71080.6, 20.0)  [250]
};
\addlegendentry{Graphite}
\addplot[scpts, sSDE] coordinates {
    (599.6, 61.5)  [1]
    (1760.5, 55.0)  [5]
    (3161.5, 45.3)  [10]
    (7386.6, 22.0)  [25]
    (14452.5, 20.5)  [50]
    (28688.9, 19.9)  [100]
    (71080.6, 19.6)  [250]
};
\addlegendentry{AMDEN}
\addplot[scpts, sShortcut] coordinates {
    (320.5, 24.4)  [1]
    (951.9, 17.8)  [5]
    (1492.3, 16.2)  [10]
    (3431.1, 17.2)  [25]
    (6436.5, 17.3)  [50]
    (11593.0, 17.5)  [100]
    (28496.9, 17.6)  [250]
};
\addlegendentry{\archname{}}
\end{axis}
\end{tikzpicture}%
}
        \vspace{-10pt}
        \caption{Young's Modulus [GPa]}
    \end{subfigure}
    \hfill
    \begin{subfigure}[b]{0.32\linewidth}
        \centering

\resizebox{\linewidth}{!}{%
\begin{tikzpicture}
\begin{axis}[
  scatteraxis,
  xmode=log, log basis x=10,
  xlabel={Generation Time per 2,000 samples (s)},
  ylabel={MAPE (\%)},
  xmin=240, xmax=130000, ymin=7, ymax=53,
  xtick={1000,10000,100000},
  ytick={10,20,30,40,50},
  legend pos=north east,
]
\addplot[scpts, sCDVAE] coordinates {
    (689.5, 50.5)  [1]
    (2024.6, 46.1)  [5]
    (3635.8, 39.0)  [10]
    (8494.6, 15.2)  [25]
    (16620.4, 14.1)  [50]
    (32992.2, 13.3)  [100]
    (81742.6, 12.8)  [250]
};
\addlegendentry{CDVAE}
\addplot[scpts, sODE] coordinates {
    (321.4, 48.1)  [1]
    (1438.4, 41.2)  [5]
    (2852.3, 25.7)  [10]
    (7091.5, 12.0)  [25]
    (14149.8, 11.5)  [50]
    (28352.4, 11.3)  [100]
    (70565.0, 11.3)  [250]
};
\addlegendentry{CrystalFlow}
\addplot[scpts, sMatterGen] coordinates {
    (779.5, 48.3)  [1]
    (2288.7, 43.5)  [5]
    (4110.0, 36.5)  [10]
    (9602.6, 13.0)  [25]
    (18788.3, 12.0)  [50]
    (37295.6, 11.6)  [100]
    (92404.7, 11.2)  [250]
};
\addlegendentry{MatterGen}
\addplot[scpts, sGraphite] coordinates {
    (599.6, 46.3)  [1]
    (1760.5, 41.2)  [5]
    (3161.5, 34.8)  [10]
    (7386.6, 11.3)  [25]
    (14452.5, 10.7)  [50]
    (28688.9, 10.4)  [100]
    (71080.6, 10.2)  [250]
};
\addlegendentry{Graphite}
\addplot[scpts, sSDE] coordinates {
    (599.6, 45.5)  [1]
    (1760.5, 40.3)  [5]
    (3161.5, 34.1)  [10]
    (7386.6, 10.8)  [25]
    (14452.5, 10.2)  [50]
    (28688.9, 10.1)  [100]
    (71080.6, 10.0)  [250]
};
\addlegendentry{AMDEN}
\addplot[scpts, sShortcut] coordinates {
    (320.5, 12.9)  [1]
    (951.9, 9.8)  [5]
    (1492.3, 9.1)  [10]
    (3431.1, 8.8)  [25]
    (6436.5, 8.9)  [50]
    (11593.0, 8.9)  [100]
    (28496.9, 9.0)  [250]
};
\addlegendentry{\archname{}}
\end{axis}
\end{tikzpicture}%
}
        \vspace{-10pt}
        \caption{Lithium Molar Concentration (\%)}
    \end{subfigure}
    \hfill
    \begin{subfigure}[b]{0.32\linewidth}
        \centering

\resizebox{\linewidth}{!}{%
\begin{tikzpicture}
\begin{axis}[
  scatteraxis,
  xmode=log, log basis x=10,
  xlabel={Generation Time per 2,000 samples (s)},
  ylabel={MAPE (\%)},
  xmin=240, xmax=130000, ymin=10, ymax=72,
  xtick={1000,10000,100000},
  ytick={10,20,30,40,50,60,70},
  legend pos=north east,
]
\addplot[scpts, sCDVAE] coordinates {
    (689.1, 66.8)  [1]
    (1999.1, 56.0)  [5]
    (3680.8, 47.8)  [10]
    (8667.4, 19.9)  [25]
    (16648.9, 18.9)  [50]
    (32854.7, 17.8)  [100]
    (81493.1, 17.4)  [250]
};
\addlegendentry{CDVAE}
\addplot[scpts, sODE] coordinates {
    (325.6, 56.0)  [1]
    (1438.7, 50.6)  [5]
    (2840.9, 38.2)  [10]
    (7083.8, 15.9)  [25]
    (14202.4, 15.5)  [50]
    (28200.6, 15.3)  [100]
    (70543.7, 15.3)  [250]
};
\addlegendentry{CrystalFlow}
\addplot[scpts, sMatterGen] coordinates {
    (778.9, 62.8)  [1]
    (2259.9, 52.4)  [5]
    (4160.9, 44.5)  [10]
    (9797.9, 17.1)  [25]
    (18820.5, 16.2)  [50]
    (37140.1, 15.5)  [100]
    (92122.6, 15.2)  [250]
};
\addlegendentry{MatterGen}
\addplot[scpts, sGraphite] coordinates {
    (599.2, 58.5)  [1]
    (1738.3, 49.8)  [5]
    (3200.7, 41.5)  [10]
    (7536.9, 14.9)  [25]
    (14477.3, 14.4)  [50]
    (28569.3, 13.9)  [100]
    (70863.6, 13.9)  [250]
};
\addlegendentry{Graphite}
\addplot[scpts, sSDE] coordinates {
    (599.2, 59.9)  [1]
    (1738.3, 49.2)  [5]
    (3200.7, 42.6)  [10]
    (7536.9, 14.2)  [25]
    (14477.3, 13.7)  [50]
    (28569.3, 13.5)  [100]
    (70863.6, 13.6)  [250]
};
\addlegendentry{AMDEN}
\addplot[scpts, sShortcut] coordinates {
    (307.4, 18.5)  [1]
    (931.7, 14.0)  [5]
    (1427.3, 13.1)  [10]
    (3302.5, 12.1)  [25]
    (6203.8, 12.0)  [50]
    (11599.6, 12.1)  [100]
    (28425.5, 12.4)  [250]
};
\addlegendentry{\archname{}}
\end{axis}
\end{tikzpicture}%
}
        \vspace{-10pt}
        \caption{Shear Modulus [GPa]}
    \end{subfigure}
    \caption{Inverse design metrics versus generation time per 2,000 MEG samples. Labels indicate the number of sampling steps.}
    \label{fig:meg-inverse-accuracy-vs-time}
\end{figure*}

\subsection{Scatter Plots of Evaluation Metrics versus Generation Time}
\label{apx:scatter-plot}
Figures~\ref{fig:structural-accuracy-vs-time} to \ref{fig:meg-inverse-accuracy-vs-time} illustrate the structural and inverse design metrics versus generation time of different models and generation step counts. These figures provide an intuitive comparison of the inference efficiency of \archname{} and the baseline models.

\begin{figure*}[t]
\centering
\begin{minipage}[b]{0.24\linewidth}\centering\resizebox{\linewidth}{!}{%
\begin{tikzpicture}
\begin{axis}[
  width=5cm, height=3cm,
  xlabel={$\sigma_{\max}^{\mX}$ (\AA)},
  ylabel={MAPE (\%)},
  xmin=0.85, xmax=2.65,
  ymin=2.5, ymax=6,
  xtick={1.0,1.5,2.0,2.5},
  ytick={3,4,5,6},
  tick label style={font=\scriptsize},
  label style={font=\small},
  grid=major, grid style={gray!20},
]
\addplot[red!70!black, thick, mark=*, mark size=1pt]
  coordinates {
    (1.0, 4.1) (1.3, 3.8) (1.5, 3.1)
    (1.7, 3.0)
    (2.0, 3.3) (2.5, 4.2)
  };
\draw[dashed, gray, thin] (axis cs:1.7, 2.5) -- (axis cs:1.7, 6);
\end{axis}
\end{tikzpicture}%
}\end{minipage}\hfill
\begin{minipage}[b]{0.24\linewidth}\centering\resizebox{\linewidth}{!}{%
\begin{tikzpicture}
\begin{axis}[
  width=5cm, height=3cm,
  xlabel={$\sigma_{\max}^{\mE}$},
  ylabel={MAPE (\%)},
  xmin=0.4, xmax=2.6,
  ymin=2.5, ymax=6,
  xtick={0.5,1.0,1.5,2.0,2.5},
  ytick={3,4,5,6},
  tick label style={font=\scriptsize},
  label style={font=\small},
  grid=major, grid style={gray!20},
]
\addplot[red!70!black, thick, mark=*, mark size=1pt]
  coordinates {
    (0.5, 4.35) (1.0, 3.22)
    (1.5, 3.0)
    (2.0, 3.8) (2.5, 5.0)
  };
\draw[dashed, gray, thin] (axis cs:1.5, 2.5) -- (axis cs:1.5, 6);
\end{axis}
\end{tikzpicture}%
}\end{minipage}\hfill
\begin{minipage}[b]{0.24\linewidth}\centering\resizebox{\linewidth}{!}{%
\begin{tikzpicture}
\begin{axis}[
  width=5cm, height=3cm,
  xlabel={$\lambda_{\mE}$},
  ylabel={MAPE (\%)},
  xmin=0, xmax=1.6,
  ymin=2.5, ymax=6,
  xtick={0.5,1.0,1.5,2.0},
  ytick={3,4,5,6},
  tick label style={font=\scriptsize},
  label style={font=\small},
  grid=major, grid style={gray!20},
]
\addplot[red!70!black, thick, mark=*, mark size=1pt]
  coordinates {
    (0.1, 4.82) (0.25, 3.46)
    (0.5, 3.0)
    (1.0, 3.7) (1.5, 4.6)
  };
\draw[dashed, gray, thin] (axis cs:0.5, 2.5) -- (axis cs:0.5, 6);
\end{axis}
\end{tikzpicture}%
}\end{minipage}\hfill
\begin{minipage}[b]{0.24\linewidth}\centering\resizebox{\linewidth}{!}{%
\begin{tikzpicture}
\begin{axis}[
  width=5cm, height=3cm,
  xlabel={$\rho_{\text{SC}}$ (\%)},
  ylabel={MAPE (\%)},
  xmin=8, xmax=52,
  ymin=2.5, ymax=6,
  xtick={10,20,30,40,50},
  ytick={3,4,5,6},
  tick label style={font=\scriptsize},
  label style={font=\small},
  grid=major, grid style={gray!20},
]
\addplot[red!70!black, thick, mark=*, mark size=1pt]
  coordinates {
    (10, 5.023) (15, 3.5) (20, 3.15)
    (25, 3.0)
    (35, 2.94) (50, 2.92)
  };
\draw[dashed, gray, thin] (axis cs:25, 2.5) -- (axis cs:25, 6);
\end{axis}
\end{tikzpicture}%
}\end{minipage}

\begin{minipage}[b]{0.24\linewidth}\centering\resizebox{\linewidth}{!}{%
\begin{tikzpicture}
\begin{axis}[
  width=5cm, height=3cm,
  xlabel={$p_{\text{drop}}$},
  ylabel={MAPE (\%)},
  xmin=0.06, xmax=0.54,
  ymin=2.5, ymax=6,
  xtick={0.1,0.2,0.3,0.4,0.5},
  ytick={3,4,5,6},
  tick label style={font=\scriptsize},
  label style={font=\small},
  grid=major, grid style={gray!20},
]
\addplot[red!70!black, thick, mark=*, mark size=1pt]
  coordinates {
    (0.1, 3.87) (0.15, 3.35)
    (0.2, 3.0)
    (0.3, 3.52) (0.4, 4.68) (0.5, 5.3)
  };
\draw[dashed, gray, thin] (axis cs:0.2, 2.5) -- (axis cs:0.2, 6);
\end{axis}
\end{tikzpicture}%
}\end{minipage}\hfill
\begin{minipage}[b]{0.24\linewidth}\centering\resizebox{\linewidth}{!}{%
\begin{tikzpicture}
\begin{axis}[
  width=5cm, height=3cm,
  xlabel={$r_c$ (\AA)},
  ylabel={MAPE (\%)},
  xmin=5.35, xmax=7.65,
  ymin=2.5, ymax=6,
  xtick={5.5,6.0,6.5,7.0,7.5},
  ytick={3,4,5,6},
  tick label style={font=\scriptsize},
  label style={font=\small},
  grid=major, grid style={gray!20},
]
\addplot[red!70!black, thick, mark=*, mark size=1pt]
  coordinates {
    (5.5, 5.4) (6.0, 3.7) (6.5, 3.0) (7.0, 3.11) (7.5, 3.26)
  };
\draw[dashed, gray, thin] (axis cs:6.5, 2.5) -- (axis cs:6.5, 6);
\end{axis}
\end{tikzpicture}%
}\end{minipage}\hfill
\begin{minipage}[b]{0.24\linewidth}\centering\resizebox{\linewidth}{!}{%
\begin{tikzpicture}
\begin{axis}[
  width=5cm, height=3cm,
  xlabel={$L$},
  ylabel={MAPE (\%)},
  xmin=0.7, xmax=6.3,
  ymin=2.5, ymax=6,
  xtick={1,2,3,4,5,6},
  ytick={3,4,5,6},
  tick label style={font=\scriptsize},
  label style={font=\small},
  grid=major, grid style={gray!20},
]
\addplot[red!70!black, thick, mark=*, mark size=1pt]
  coordinates {
    (1, 5.4) (2, 3.58)
    (3, 3.0)
    (4, 2.94) (5, 3.02) (6, 3.08)
  };
\draw[dashed, gray, thin] (axis cs:3, 2.5) -- (axis cs:3, 6);
\end{axis}
\end{tikzpicture}%
}\end{minipage}\hfill
\begin{minipage}[b]{0.24\linewidth}\centering\resizebox{\linewidth}{!}{%
\begin{tikzpicture}
\begin{axis}[
  width=5cm, height=3cm,
  xlabel={$d_h$},
  ylabel={MAPE (\%)},
  xmode=log, log basis x=2,
  xmin=13, xmax=310,
  ymin=2.5, ymax=6,
  xtick={16,32,64,128,256},
  xticklabels={16,32,64,128,256},
  ytick={3,4,5,6},
  tick label style={font=\scriptsize},
  label style={font=\small},
  grid=major, grid style={gray!20},
]
\addplot[red!70!black, thick, mark=*, mark size=1pt]
  coordinates {
    (16, 5.22) (32, 3.84)
    (64, 3.0)
    (128, 2.95) (256, 2.93)
  };
\draw[dashed, gray, thin] (axis cs:64, 2.5) -- (axis cs:64, 6);
\end{axis}
\end{tikzpicture}%
}\end{minipage}
\caption{Sensitivity sweeps of \archname{} hyperparameters. Dashed lines mark the chosen values.}
\label{fig:hp-sweeps}
\end{figure*}

\begin{figure*}[!t]
    \centering
    \begin{subfigure}[b]{0.19\linewidth}
        \centering
        \includegraphics[width=\linewidth]{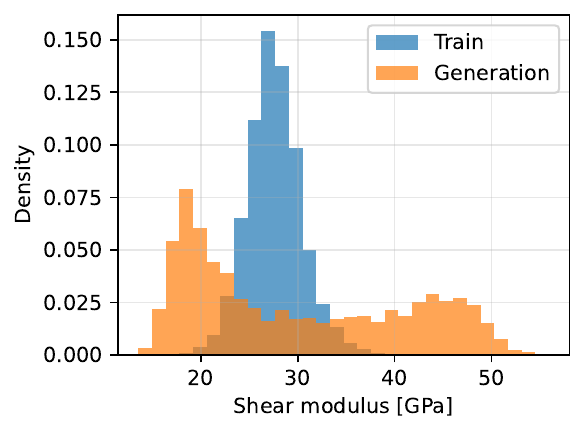}
        \caption{$G$ of a-\ch{SiO2} samples}
    \end{subfigure}
    \hfill
    \begin{subfigure}[b]{0.19\linewidth}
        \centering
        \includegraphics[width=\linewidth]{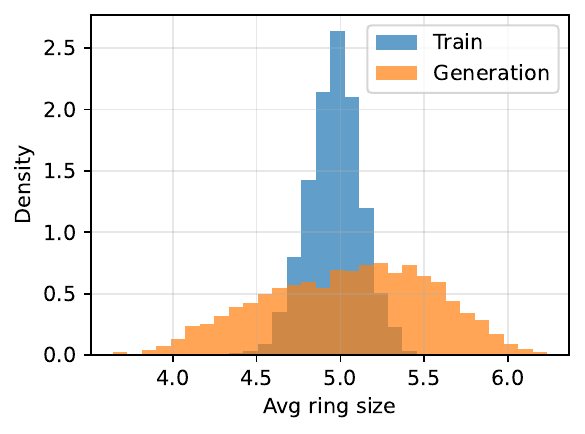}
        \caption{RSD of a-\ch{SiO2} samples}
    \end{subfigure}
    \hfill
    \begin{subfigure}[b]{0.19\linewidth}
        \centering
        \includegraphics[width=\linewidth]{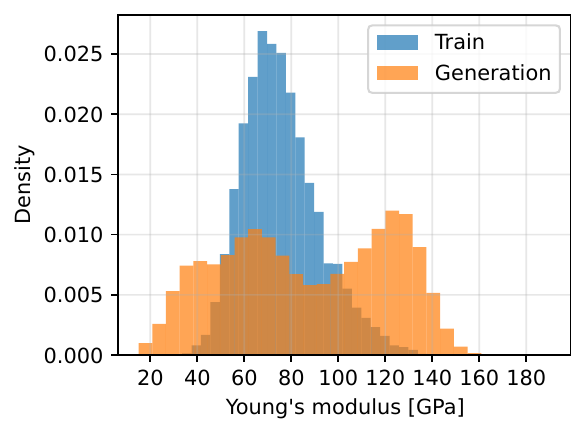}
        \caption{$E$ of MEG samples}
    \end{subfigure}
    \hfill
    \begin{subfigure}[b]{0.19\linewidth}
        \centering
        \includegraphics[width=\linewidth]{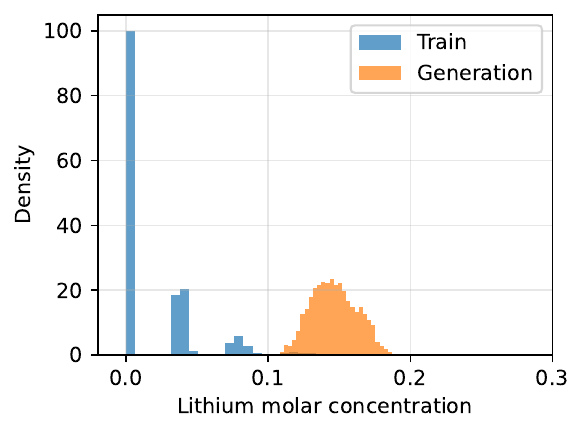}
        \caption{$C_\text{Li}$ of MEG samples}
    \end{subfigure}
    \hfill
    \begin{subfigure}[b]{0.19\linewidth}
        \centering
        \includegraphics[width=\linewidth]{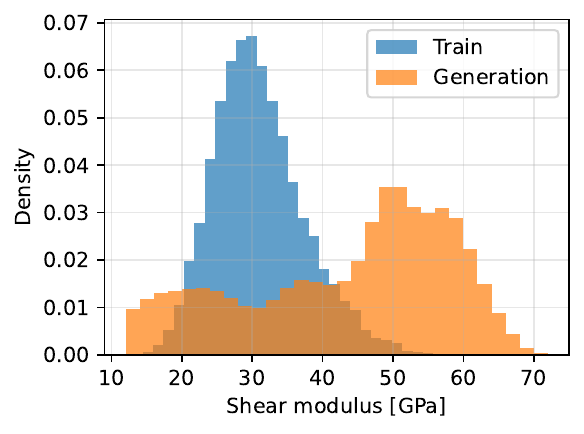}
        \caption{$G$ of MEG samples}
    \end{subfigure}
    \caption{Distribution comparison of properties in the training samples versus the samples generated by \archname{}.}
    \label{fig:property-histograms}
\end{figure*}

\subsection{Hyperparameter Sensitivity}
\label{apx:hp-sensitivity}
We verify that the critical hyperparameters of \archname{} are set at their optimal values through sensitivity sweeps.
Figure~\ref{fig:hp-sweeps} sweeps eight hyperparameters that control the core components of \archname, each evaluated by the a-\ch{SiO2} RSD MAPE at $n_s=10$ sampling steps.

The two noise scales set how far the material SDE perturbs the positions and the element embeddings.
The position noise scale $\sigma_{\max}^{\mX}$ controls the positions.
A value below $1.7$\,\AA{} injects too little noise to reach disordered atomic configurations.
A value above $1.7$\,\AA{} moves the positions so far from any physical structure that the denoiser cannot recover the fine geometry within ten steps.
The chosen $1.7$\,\AA{} covers the spread of first and second neighbor distances in amorphous silica while keeping local order recoverable.
The element noise scale $\sigma_{\max}^{\mE}$ controls the element embeddings.
Too small a value barely perturbs the element embeddings, so the model never learns to correct atom types during sampling.
Too large a value washes out the embedding of each element, so the denoiser loses the atom type it must recover.
The chosen $1.5$ perturbs atom types enough to learn meaningful reassignments while keeping each atom type recoverable.

The element loss weight $\lambda_{\mE}$ balances the position and element objectives in training.
A small $\lambda_{\mE}$ gives the element prediction too weak a signal, while a large $\lambda_{\mE}$ starves the position objective.
The chosen $0.5$ gives the element loss enough weight to fix the atom types while keeping the position loss dominant enough to place atoms accurately.

The shortcut-loss batch fraction $\rho_\text{SC}$ sets how many training batches carry the shortcut self-consistency loss that enables generation in few sampling steps.
At a small $\rho_\text{SC}$ the self-consistency objective sees too few batches, making \archname{} poorly trained.
The error drops as $\rho_\text{SC}$ grows and flattens near $25\%$. Beyond $25\%$, the error sees little improvement, while the training cost grows.
The chosen $\rho_\text{SC}=25\%$ keeps the accuracy at few sampling steps while cutting the shortcut training cost, which supports the training efficiency of \archname.

The property dropout probability $p_{\text{drop}}$ sets how often each property is dropped during training, which teaches the flexible property embedding to condition on different subsets of properties.
A small $p_{\text{drop}}$ almost always shows the model the full property set, making subset conditioning generalize poorly at few sampling steps.
A large $p_{\text{drop}}$ rarely shows the full property set, so conditioning on all properties together weakens.
The chosen $0.2$ gives balanced exposure to every subset size.

The last three sweeps size the E(n)-equivariant backbone.
The neighbor cutoff radius $r_c$ sets the neighbor graph.
Below $6.5$\,\AA{} the graph is truncated before the second neighbor shell, so message passing misses the medium-range order.
Above $6.5$\,\AA{} the graph pulls in distant atoms that add noise and inflate its density.
The chosen $6.5$\,\AA{} spans the first and second neighbor shells.
The number of layers $L$ and the hidden dimension $d_h$ set the depth and width of the backbone.
With one or two layers the backbone cannot span the medium-range order and underfits, and a hidden dimension of $16$ or $32$ is too narrow to represent the local atomic environments.
Beyond $L=3$ and $d_h=64$ the accuracy saturates while the computational cost keeps growing.
The chosen values give the best balance between efficiency and expressiveness of the backbone.

\subsection{Property Distributions of Generated Samples}
\label{apx:property-histograms}
Figure~\ref{fig:property-histograms} compares the property distributions of the training samples with those of the samples generated by \archname{} with 100 sampling steps, on the a-\ch{SiO2} and MEG datasets.
The target properties given to the model extend beyond the training distribution, and the generated samples follow these targets, which demonstrates the extrapolation capability of \archname.

\section{Discussion}
\label{apx:discussion}

\paragraph{Limitations.}
Recent work~\citep{finkler2025inverse} discovers that diffusion models face inherent limitations in generating annealed structures, i.e., structures cooled down slowly and with lower energy compared to liquid-state ones. This limitation cannot be overcome by using more sampling steps, but can be overcome by incorporating physics-guided Hamiltonian Monte Carlo (HMC) refinement, with the downside of a slow sampling process. This process cannot be accelerated by learning shortcuts, since HMC refinement is a search-based solution, while diffusion model sampling is prediction-based. Thus, solutions for improving sampling efficiency on annealed structures require further exploration.

\paragraph{Scalability and validation.}
\archname{} constructs the atomic graph with a fixed cutoff radius, so the number of edges grows linearly with the number of atoms.
Generation therefore scales linearly in system size, and handling cells with thousands of atoms is limited by GPU memory rather than by the method itself.
The three datasets are chosen to span single-element (a-\ch{Si}), binary (a-\ch{SiO2}), and 11-element (MEG) systems, which are representative of realistic amorphous materials.
The shortage of large-scale, multi-element amorphous datasets with property labels remains a central open challenge of the field~\citep{finkler2025inverse}.
We follow the standard protocol of validating inverse design in simulation, since obtaining real samples that match arbitrary target properties or exact 3D structures is generally infeasible.

\end{document}